\documentclass[journal]{IEEEtran}
\usepackage{todonotes}
\usepackage{mathtools}
\usepackage{bbm}
\usepackage{amsmath} 
\usepackage{bm}
\usepackage{mathtools}
\usepackage[acronym]{glossaries}
\usepackage{array}
\usepackage{graphicx}
\usepackage{caption}
\usepackage{subcaption}
\usepackage{multirow}
\usepackage[normalem]{ulem}
\usepackage[makeroom]{cancel}

\newacronym{dropall}{\textit{All-Dropout}}{\textit{All-Dropout}}
\newacronym{allconv}{\textit{All-Convolutional}}{\textit{All-Convolutional}}
\newacronym{invertednet}{\textit{InvertedNet}}{\textit{InvertedNet}}

\newcommand{\udropall}{All-Dropout}
\newcommand{\uallconv}{All-Convolutional}
\newcommand{\invertednet}{InvertedNet}
\newcommand{\originalunet}{Original U-Net}

\usepackage[noadjust]{cite}

\ifCLASSINFOpdf
\else
\fi

\begin{document}
%
\title{Fully Convolutional Architectures for Multi-Class Segmentation in Chest Radiographs}
%

\author{Alexey~A.~Novikov,~
	Dimitrios~Lenis,~
	David~Major,~
	Ji{\v r}{\'{\i}} Hlad\r{u}vka,~
	Maria~Wimmer,~
	and~Katja~B\"{u}hler
	\thanks{A. A. Novikov, D. Lenis, D. Major,  J. Hlad\r{u}vka, M. Wimmer and K. B\"{u}hler are with the VRVis Center for Virtual Reality and Visualization, 1220 Vienna, Austria, e-mail: (novikov@vrvis.at, lenis@vrvis.at, major@vrvis.at,  hladuvka@vrvis.at, mwimmer@vrvis.at, buehler@vrvis.at).}
}

%
%

\markboth{}%
{Shell \MakeLowercase{\textit{et al.}}: Bare Demo of IEEEtran.cls for IEEE Journals}
%



\maketitle


\begin{abstract}

The success of deep convolutional neural networks on image classification and recognition tasks has led to new applications in very diversified contexts, including the field of medical imaging. In this paper we investigate and propose neural network architectures for automated multi-class segmentation of anatomical organs in chest radiographs, namely for lungs, clavicles and heart. 

We address several open challenges including model overfitting, reducing number of parameters and handling of severely imbalanced data in CXR by fusing recent concepts in convolutional networks and adapting them to the segmentation problem task in CXR. We demonstrate that our architecture combining delayed subsampling, exponential linear units, highly restrictive regularization and a large number of high resolution low level abstract features outperforms state-of-the-art methods on all considered organs, as well as the human observer on lungs and heart.

The models use a multi-class configuration with three target classes and are trained and tested on the publicly available JSRT database, consisting of 247 X-ray images the ground-truth masks for which are available in the SCR database. Our best performing model, trained with the loss function based on the Dice coefficient, reached mean Jaccard overlap scores of 95.0\% for lungs,  86.8\% for clavicles and 88.2\% for heart. This architecture outperformed the human observer results for lungs and heart.

\end{abstract}


%
\IEEEpeerreviewmaketitle

\section{Introduction}

%
%
%
%

\IEEEPARstart{D}{espite} a plethora of modalities and their combinations in current state-of-the-art medical imaging, radiography holds an esteemed position, forming together with ultrasonography the two main pillars of diagnostic imaging, helping in solving between 70-80\% of diagnostic questions~\cite{WHO}. The importance of chest radiographs (CXR) is particularly clear: they are the most common images taken in medicine and therefore one of the leading tools in diagnosis and treatment.
 
Segmentation of anatomical structures in CXR plays an important role in many tasks involved in computer-aided diagnosis. In particular, certain diagnostic procedures directly benefit from accurate boundaries of lung fields. For example, irregular shape, size measurements and total lung area can give clues to early manifestations of serious diseases, including cardiomegaly, emphysema and many others. Clavicle segmentation is crucial, especially for early diagnosis because the lung apex is the area where tuberculosis and many other lung diseases most commonly manifest \cite{Ginneken2002}. Heart segmentation is often used to measure the cardiothoracic ratio (abnormally large heart) with the aim to detect cardiomegaly \cite{Nakamori1990}.


CXRs are challenging for many tasks related to medical image analysis.  Individual anatomical intricacies such as high interpersonal variations in shape and size of central organs, due to age, size and gender, combine with ambiguous organ boundaries resulting from organ overlaps and artifacts caused by movements and image modality intrinsics. For instance consider the overlaps of lung fields with clavicles and the rib cage: lung border intensity is not always consistent, especially near the heart area, which leads to fuzzy intensity transitions.
Furthermore, the upper and lower part of the heart boundaries are not clearly visible. While these issues cause even experienced radiologists trouble in consistent interpretation of CXR images~\cite{Quekel1999}, algorithmic approaches face additional problems. One of these problems for CXR is the unequal distribution between its classes. In particular, lung class severely outrepresents clavicle class in terms of pixel area. As an example, in the SCR (Segmentation in Chest Radiographs) database \cite{Ginneken2006}, containing manual segmentations for lung fields, heart and clavicles, for the JSRT (Japanese Society of Radiological Technology) image set \cite{Shiraishi2000}, 73.53\% of pixels belong to lungs, 4.62\% to clavicles and 21.85\% to hearts. This demonstrates a severe between-class imbalance in the data. Consequently, multi-class segmentation in CXR still remains a challenging problem. 

In this paper, we evaluate three different fully-convolutional architectures and introduce the \invertednet \, as the best performing based on statistical tests and Jaccard overlap scores on the test set. InvertedNet is built on the idea of delayed subsampling in the contraction part of the network and large number of high resolution low level abstract features. We also propose to use exponential linear units and train the network with a normalized class frequency aware loss function based on the Dice coefficient. InvertedNet has ten times less parameters compared to the U-Net but still preserves the accuracy and achieves better results for the segmentation of all considered organs, with the largest statistically significant improvements achieved for clavicles.

This paper is organized as follows. Sec.~\ref{sub_sec:related} gives a short introduction to the related work on lung, heart and clavicle segmentation in CXR. Sec.~\ref{sec:methodology} presents the methodology of our approach. Sec.~\ref{sec:exp_setup} describes the experimental setup of our evaluations and Sec.~\ref{sec:results} presents and discusses the results. Sec.~\ref{sec:conclusions} concludes the paper and discusses future work. 


\subsection{Related Work}
\label{sub_sec:related}

\begin{figure*}[th!]
	\centering
	\includegraphics[width=0.9\textwidth]{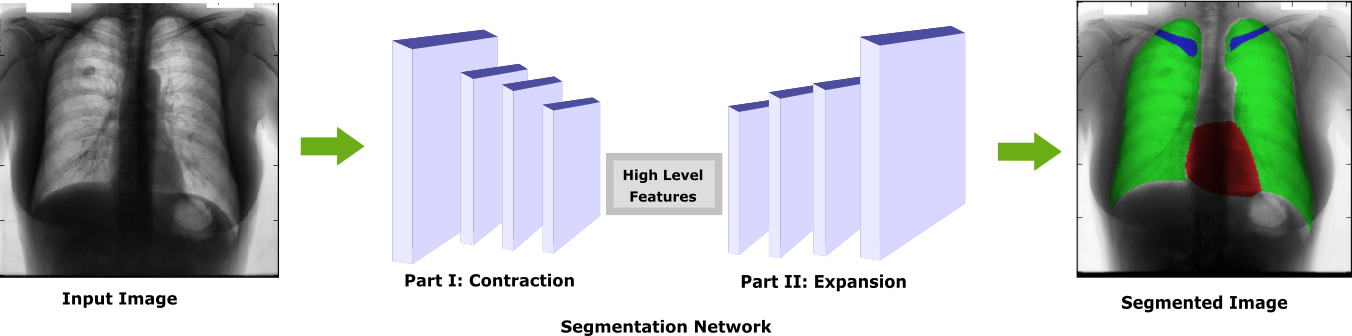}
	\caption{Multi-class segmentation process: from input chest CXR to three masks of lungs, clavicles and heart}
	\label{fig:pipeline}
\end{figure*}

Semantic CXR segmentation has been studied extensively in the literature. Unsurprisingly, solutions therefore vary in their favoured toolsets and targets. Along with the introduction of the SCR database, van Ginneken et al.~\cite{Ginneken2006} compared performance results of selected methods. In the following we mainly focus on solutions that have been tested against this common ground-truth set.

Note that the listed methodologies were performed on down-sampled images of smaller resolution, namely $256\times256$. While adding problematic border smoothing, this scale allows a reduction in computational complexity and makes some approaches tractable.  In accordance with the aforementioned study, we use the widely recognized Jaccard Index \cite{Tanimoto1958} overlap measure as our standard of comparison.


\subsubsection*{Classical Approaches}
Following and augmenting van~Ginneken~et~al.~\cite{Ginneken2006}, the space of algorithmic approaches may be roughly partitioned into \emph{rule-}, \emph{shape-} and \emph{graph-}based methods, \emph{pixel classification} and \emph{statistical approaches}. Each methodological framework has its own set of advantages. By limiting to a predefined ruleset or deformable shape, rule- and shape-based methods will yield anatomically sound solutions. Graph-based methods build upon the anatomically inherent topology and therefore will also adhere to that principle, while simultaneously allowing for a higher class of variations in exchange for higher computational complexity. Pixel classification and statistical approaches treat the problem as a local classification / optimization task and therefore allow for higher variation, maintaining traceable computation at the cost of sometimes unrealistic outcomes. Typically, as the following shows, better results were achieved in many cases via hybrid approaches, such as solutions combining efficient initial segmentations with successive detailed adjustments to plausible outputs.

\emph{Lung fields} have received significant attention. 
A recent hybrid approach stems from the work by Shao et al.~\cite{Shao2014}, combining active shape and appearance models; the group produced a Jaccard score of $0.946$. Ibragimov et al.~\cite{Ibragimov2016} demonstrated how the combination of landmark-based segmentation and a random forest classifier can achieve an overlap score of $0.953$, while still maintaining computational efficiency. An active shape model approach by Xu et al.~\cite{Xu2012}, specifically addressing the initialization dependency of these models, achieves an improvement and an overlay score of $0.954$. Seghers et al.~\cite{Seghers2007} used dynamic programming to optimize the loss function consisting of shape and intensity models. The score of $0.951$ was achieved. Van Ginneken et al.~\cite{Ginneken2006} survey older approaches, back to 2006, that score between $0.713$ and $0.949$ on the same dataset; their survey also evaluated a human observer with a score of $0.946$, which did not exhibit statistically significant variation from the survey leading pixel classification method. Overall, lung field segmentation in CXR remains an active topic, with algorithmic approaches rivaling human observer performance.

In comparison, segmentation of \emph{clavicles} has proven more challenging. High variations in their positioning and general shape, the impact of bone density on the radiograph and their overlap with rib and lung structures result in interpatient anatomical variations, leading to major deviations from a theoretical average clavicle and hence a steep impact on overlap scores. Van Ginneken et al.~\cite{Ginneken2006} include clavicle segmentation in their survey where the compared methods were able to reach scores between $0.505$ and $0.734$. The human observer reached $0.896$, which demonstrates that, compared to segmentation of lung and heart fields, this task is challenging even for humans. Hogeweg et al.~\cite{Hogeweg2012} developed a combination of pixel classification, active shape model and dynamic programming that led them to an overlap of $0.860$; however, this overlap was only measured within the lung fields. Predominantly shape/contour- based models vary on the choice of the underlying feature space. Exemplified in the approach presented by Boussaid et al.~\cite{Boussaid2014}, the problem is addressed as a deformable contour model that uses SIFT features to describe the embedding object appearance.  A mean overlap score of $0.801$ was reported.

Starting from their respective lung field segmentation, most approaches are geared towards a generalizable solution, trying to adapt their algorithm to the \emph{heart} segmentation task. Similarly, van Ginneken et al.~\cite{Ginneken2006} also report on the segmentation results of different approaches, with mean overlap scores between $0.77$ and $0.86$. Boussaid et al.~\cite{Boussaid2014} use a machine learning approach with deformable contour models to improve shape detection and report mean overlap score of $0.91$. 
Generally, as the heart boundaries are overlapped and occluded by the surrounding lung fields, and hence not clearly visible, exact segmentation remains challenging.

\subsubsection*{Neural networks}
While conceptually more than 50 years old, neural networks (NNs), the abstracted basis of \emph{deep learning}, are undergoing a revival \cite{LeCun2015}. A deeper understanding of training and numerical behavior and the steep increase in tractable calculation schemes by leveraging graphical processing units (GPUs) has allowed this class of approach to become the \emph{de facto} standard, or at least a serious contender in several machine learning branches \cite{LeCun2015, Lai2015}. The following focuses on convolutional neural networks (CNNs), as the dominant subclass of NN in computer vision tasks~\cite{LeCun2015}. A prototypical setup for such CNNs consists of a combination of convolution filters, interspersed with  pooling layers. The driving idea is to mimic human visual cognition, in the sense that the complete picture is derived from low-level features, e.g., edges and circles, which in return yield more distinctive features and finally the desired target through recombination in each successive layer. 

For segmentation of medical images several such setups have been studied; e.g., Greenspan et al.~\cite{IEEE2016} made a summary of the recent state-of-the-art works in the field. The \emph{semantic segmentation} typically builds upon a vast set of training data, e.g., Pascal VOC-2012~\cite{Chen2014}. Such large datasets are not typical for the medical domain. This renders most current approaches unfeasible, hence calling for a finely tailored strategy. 

Long et al.~\cite{Long2015} introduced the concept of the \emph{Fully~Convolutional~Net}. This type of network takes an input of arbitrary size and produces correspondingly-sized output with efficient inference. In combination with layer fusion, i.e., shortcuts between selected layers, this configuration achieves a nonlinear, local-to-global feature representation, and allows for a pixelwise classification. By adapting this network class with successive upsampling layers, i.e., enlarging the field of view of the convolution, Ronneberger et al. \cite{Ronneberger2015} guide the resolution of feature extraction, and thereby control the local-to-global relations of features. The proposed U-Net consists of contraction and expansion parts: in the contraction part high-level abstract features are extracted by consecutive application of pairs of convolutional and pooling layers, while in the expansion part the low-level abstract features are merged with the features from the contractive part. The U-Net has recently been evaluated in the CXR segmentation task~\cite{Wang2017}. 

The idea of enhancing and guiding information flow through this kind of networks has also been a driving idea of current general image recognition competition leading networks such as ResNet \cite{ResNet2016} that gave inspiration to a series of solutions and ensembles of them \cite{DenseNet2016,ImageNetWinners2017}. 

Motivated by its performance on other segmentation tasks reported in the literature, and inspired by the ideas of flexibility and efficiency of information flows in the segmentation networks, we adapted the U-Net specifically for CXRs by adding modifications to its \textit{architecture} and introducing a different \textit{training} strategy, with the aim of improving the results on the challenging tasks of clavicle and heart segmentation.

\subsection{Contributions}

\begin{itemize}
	\item We propose a multi-class approach for segmentation of X-ray images, and show its applicability on CXRs particularly for lung fields, clavicles, and hearts. This approach can also potentially be used for segmentation of other anatomical organs and other types of radiograms. 
	\item We show that the loss function based on the Dice coefficient allows higher overlap scores than when trained with the loss function based on the cross entropy coefficient. 
	\item We show that using exponential linear units (ELUs) \cite{Clevert2015} instead of rectified linear units (ReLUs) \cite{Hahnloser2000} can speed up training and achieve higher overlap scores.  
	\item Compared to the U-Net, our best architecture employs around ten times fewer parameters, while preserving the accuracy and achieving better results for the segmentation of all considered organs, with the largest improvements achieved for clavicles. 
	\item Our best architecture yields overlap scores comparable and in many cases surpassing state-of-the-art techniques and human performance on all considered tasks, including the classically challenging cases of heart and clavicle segmentation. By performing Wilcoxon signed-rank tests on the segmentation results we show that the difference in performance between our proposed models and original U-Net is statistically significant. 
	\item To the best of our knowledge, our solution is the fastest multi-class CXR segmentation approach to date, allowing very efficient processing of large amounts of data.   
\end{itemize}



\section{Methodology}
\label{sec:methodology}


In this section, we begin with a formal description of the multi-class approach (Sec.~\ref{sub_sec:multilabel_approach}). 
Our training strategies and proposed architectures for CXR images are described in Sec.~\ref{sub_sec:architectures_proposed}. 

\subsection{Multi-Class Approach}
\label{sub_sec:multilabel_approach}

The dataset used for our experiments is a set of $n$ images $\mathcal{I}=\{I_1, ... , I_n \}$ with pixels $\bm{x} = (x_1, x_2)$ on a discrete grid $m_1 \times m_2 $ and intensities $I_i(\bm{x}) \in \mathcal{J} \subset \mathbf{R}$. Additionally, for each image $I \in \mathcal{I}$, a sequence $M_I \coloneqq (M_{I,\,l})_{l=1} ^m$  of ground-truth masks is available, where $l$ corresponds to the sequence of $m$ semantic class lables $\mathcal{L} = \{l_1, ..., l_m \}$ and $M_l \in \mathcal{M}~\coloneqq~M^{m_1 \times m_2 } ( \{0,1\} ) $, the space of all binary matrices of size  $m_1 \times m_2$. 

Let  $\mathcal{M}^\prime  \coloneqq  M^{m_1 \times m_2 } ( \{0,..., | \mathcal{L} |\} )$ denote the space of matrices of size $m_1 \times m_2$ with values $\{0,..., | \mathcal{L} |\}$ that correspond to the semantic labels of $ \mathcal{L}$. Assuming each pixel of $I$ may only belong to exactly one class, $g: \mathcal{M}  \rightarrow  \mathcal{M}^\prime $, $g(M_{I}) = \sum\limits_{l=1}^{|\mathcal{L}|} l \cdot  M_{I,\,l} $ maps a given ground-truth sequence $M_{I}$ into $\mathcal{M}^\prime$. For a given image $I \in \mathcal{I}$, let $G_I \in  \mathcal{M}^\prime$ be its ground-truth matrix, derived as described above, and $\pi_l: \mathcal{M}^\prime  \rightarrow  \mathcal{M}$, $\pi_l(G_I) = M_{I,\,l} $ the projection of $G_I $ onto $\mathcal{M}$ for the semantic class $l \in \mathcal{L}$, i.e., the well-defined inverse mapping for the above operation $g$.

For training and evaluation purposes, the dataset $\mathcal{I}$ is split into three non-overlapping sets, namely  $\mathbf{I}_{\, \text{TRAIN}}$, $\mathbf{I}_{\, \text{VALID}}$ and $\mathbf{I}_{\, \text{TEST}}$. 

During training, the network is consecutively passed with minibatches $\mathcal{K} \in \mathcal{N}$, where $\mathcal{N}$ is a complete partition of the set $\mathbf{I}_{\, \text{TRAIN}}$. 
For later reference, let $c_{\mathcal{K}}$ be the total pixel count over all images $I \in \mathcal{K}$. 

For each image $I \in \mathcal{K}$, the multi-class output of the network is calculated: understanding the network as a function 
\begin{equation}
\mathcal{F} : \mathcal{I} \rightarrow \mathcal{M}^\prime,  
\end{equation}
$\mathcal{F}(I)$ derives for each pixel  $\bm{x}$ of $I$ its semantic class $l \in \mathcal{L}$ in a single step with some probability. 

In order to estimate and maximize this probability, we define a loss function
\begin{equation}
\Lambda :  \mathcal{I} \times \mathcal{M}^\prime \rightarrow \mathbf{R}
\end{equation}
that estimates the deviation (error) of the network outcome from the desired ground-truth. This error is subsequently used to update the networks parameters. This procedure is repeated until a defined set of stopping criteria are fulfilled. 

At testing time, all unseen images $I \in \mathbf{I}_{\, \text{TEST}}$ are passed through the network, their multi-label output  $\mathcal{F}(I)$ is computed and used to derive an accuracy estimate over the complete set  $\mathbf{I}_{\, \text{TEST}}$. The output channels for chest radiographs correspond to different body organs, namely clavicles, heart and lung fields. 





\subsection{Improvements to U-Net Model for Chest Radiographs}
\label{sub_sec:architectures_proposed}

Striking the balance between a neural networks size and the available training data is an ongoing challenge. Overfitting due to a vast parameter space on small training sets is probable if no special care is taken \cite{Goodfellow-et-al-2016}. While the original U-Net yields promising results in CXR segmentation, finer scrutiny showed potential for further tuning and more effective training. The number of feature maps of the unaltered network is large, increasing along its contraction part, which results in tens of millions of parameters. While in fact fully-convolutional networks can often be trained well on small sets \cite{Ronneberger2015}, our visual inspection of derived features for the U-Net trained using CXR images from the JSRT set indicate that this property might be domain specific. 
\begin{figure}[h!]
	\includegraphics[width=0.49\textwidth]{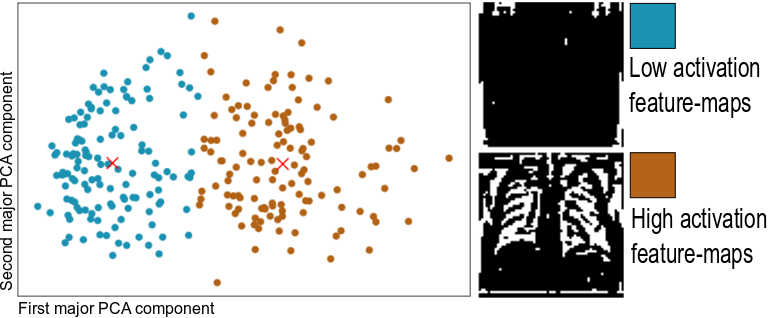}
	\caption{Feature maps grouped by their activation level. Visualizations of the low and high activation feature maps (right) correspond to the cluster centers (left). The feature maps have been thresholded at the value of 0.1 for a better visual representation.}
	\label{fig:cluster}
\end{figure}

Fig.~\ref{fig:cluster} illustrates that the network exhibits a decreasing feature map depth dependence. We grouped the convolution kernels before the third downscaling step by their influence on activations for a random test image. Out of the lower contributing kernels we randomly deactivated $25\%$ and measured the difference between the altered and unaltered network outputs by their overlap, averaged over ten runs. We repeated this procedure in a coarser setup after the last downsampling step, where we tripled the size of filter deactivations without any clustering performed beforehand. In both cases no significant difference occurred, i.e. an overlap score of $1.0$ was achieved. This indicates that network architecture and training strategy hold potential for adaptation and tuning to the domains intrinsic needs.
The above-described dichotomy of feature masks was derived by k-Means clustering on the feature maps major PCA components. The overlap of the derived segmentation masks was compared using the Jaccard score.

\begin{figure*}[th!]
	\centering
	\includegraphics[width=0.8\textwidth]{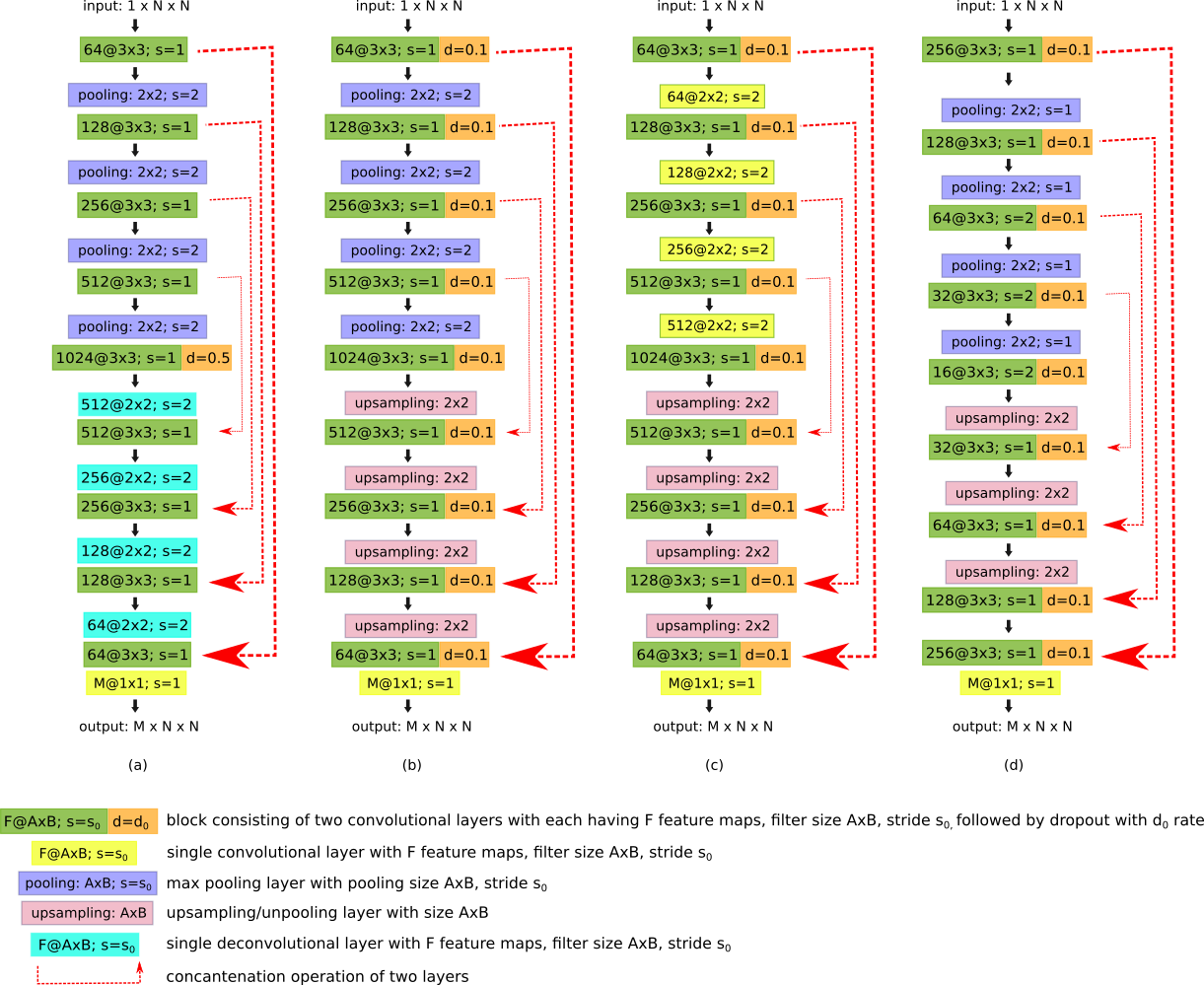}
	\caption{Overview of the evaluated architectures a) \originalunet \, b) \udropall \, c) \uallconv  \, d) \invertednet }
	\label{fig:networks}
\end{figure*}

In the following we consider a number of modifications to the U-Net to target the above-described issues. In addition to that, following the trends of simplifying the network architecture building and parameter tuning, we consider a concept of all-convolutional networks by Springenberg et al.~\cite{Springenberg2014} and propose an all-convolutional modification of the original U-Net. A schematic overview of the evaluated architectures is depicted in Fig.~\ref{fig:networks}.\\ 



\textbf{Architectural Modifications}: \\



\paragraph{\textbf{All-Dropout (fully-convolutional network with restrictive regularization)}}

In the case when the network architecture is deep and availability of training data is limited, one of the possibilities to decrease the generalization test error of the algorithm is a more restrictive regularization. As understood here, regularization is any modification intended to improve the performance of the learning algorithm on the test set. 

Due to its efficiency, adding a dropout layer \cite{Srivastava2014} has become a common practice in modern deep network architectures. 
We therefore propose and evaluate an architecture with a dropout layer placed after every convolutional layer after the activation. We use the Gaussian dropout, which is equivalent to adding a Gaussian distributed random variable with zero mean and standard deviation defined as follows:
\begin{equation*}
\sigma = \sqrt{\frac{d}{1.0-d}}
\end{equation*}
where $d \in [0,1]$ is the drop probability. 

According to Srivastava et al. \cite{Srivastava2014}, this performs even better than the classic approach which drops the neural units following a Bernoulli distribution. Besides, adding Gaussian noise is a more natural choice for CXR scans due to the occurence of this type of noise during the image acquisition~\cite{Gravel2004}. In the following evaluations, we call this modification \acrshort{dropall}. Its architecture is depicted in Fig.~\ref{fig:networks}b. \\



\paragraph{\textbf{All-Convolutional (simplifying fully-convolutional network by learning pooling)}}

Springenberg et al. \cite{Springenberg2014} made an attempt to address the issue of simplifying the convolutional networks via replacing the pooling with strided convolutions. This modification introduces new parameters into the network that can be considered as learning of pooling for each part of the network rather than just fixing pooling parameters to constant values. They showed that replacing pooling layers with convolutional layers with a higher stride or removing pooling layers completely can improve final results. Motivated by their work in order to answer if this modification can be beneficial for the fully-convolutional networks, we adapt the \acrshort{dropall} in accordance with their best performing model, All-CNN-C: each pooling layer is replaced by a convolutional layer with filter size equal to the pooling size of the replaced pooling layer. In the following evaluations, we call this architecture \acrshort{allconv}. The architecture is depicted in Fig.~\ref{fig:networks}c. \\

\paragraph{\textbf{InvertedNet (fully-convolutional network with fewer parameters)}}

It is well known that convolutional networks in general train better when regularization is used. However, even when it is used like in the case of the proposed \udropall \, with dropout placed after every convolutional layer, the number of features in the end of the contraction part is large and based on the visual inspection of the layer activations most of those do not learn very meaningful features.

Another way of dealing with model overfitting is to reduce the expressivity of the function. Motivated by the optimizations proposed by He et al.~\cite{He2015}, we propose altering the \acrshort{dropall} architecture by introducing the delayed subsampling of the pooling layers. On contrary to the work by He et al.~\cite{He2015}, on the convolutional layer after the first pooling we set the stride = 1. On all following pooling layers we set the stride = 1 and stride = 2 on their subsequent convolutional layers. 

In addition, we reduce the solution space of the network by reordering the number of feature maps in the convolutional layers. In contrast to \acrshort{dropall}, we propose starting with a large number of feature maps and dividing this by a factor of two after every pooling layer in the contraction part of the network, while increasing by a factor of two after every upsampling layer in the expansion part. 
Intuitively, such architecture seems more reasonable for more rigid anatomical organs such as clavicles, because their topologies do not vary much, meaning there is no need to learn many low resolution high level abstract features. In the following evaluation we will call this architecture \acrshort{invertednet}, due to the way the numbers of feature maps are changed with respect to the \acrshort{dropall} \, architecture. Due to a slightly unsatisfactory performance of the \uallconv \, network during initial evaluations, we kept the pooling in this architecture. We included both models into final evaluations. This architecture is depicted schematically in Fig.~\ref{fig:networks}d.\\

\textbf{Training Strategies}: \\

As mentioned above, large differences in sizes between anatomical organs of interest can introduce a problem of imbalanced data representation. In such cases, classes are represented by significantly different numbers of pixels, and therefore the learning algorithm can become biased towards the dominating class \cite{He2009}. We address the imbalance in pixel representation by introducing class weights into the loss function based on cross entropy and Dice coefficients. 

Following section \ref{sub_sec:multilabel_approach}, let $\mathcal{L}$ be the set of all ground-truth classes, $\mathcal{N}$ a complete partition of our training set $\mathbf{I}_{\, \text{TRAIN}}$, $\mathcal{K} \in \mathcal{N}$ a set of images and $c_{\mathcal{K}}$ its total pixel count. With the aim of balancing the different organ sizes, and thereby their contribution to our loss function, we introduce weighting coefficients $r_{\mathcal{K}, l}$ for each semantic class $l \in \mathcal{L}$ by the ratio: 
\begin{equation}
r_{\mathcal{K}, l} \coloneqq \frac{c_{l, \mathcal{K}}}{c_{\mathcal{K}}} 
\end{equation}
where $c_{l, \mathcal{K}}$ is the number of pixels belonging to class $l$ in the training batch $\mathcal{K}$. \\

For a distance function $d:\mathcal{I}  \times \mathcal{M}^\prime  \rightarrow \mathbf{R}$ and an image $I \in \mathcal{K}$ we define and minimize our loss function
\begin{equation}
\Lambda(I, G_I) \coloneqq -\sum_{l \in \mathcal{L}} r_{\mathcal{K}, l}^{-1} \,\, d_l( I,  G_I ) 
\end{equation}
over the set $\mathcal{K}$ and the complete partition. In this way, sparsely represented classes such as clavicles are no longer under represented in favor of, e.g., lung fields, for which ground-truth masks are severely larger in terms of pixel area. 

For $d$, we chose and evaluated the \textit{cross entropy} and \textit{Dice} distance functions. Loss functions based on cross entropy are a typical choice for neural networks, while using Dice as an overlap measure seems a natural choice for segmentation problems. 


These two functions differ in their domain, while classically \emph{Dice} distance works with binary masks, \emph{cross entropy} maps probability distributions to real numbers. Hence the final output of the network architecture must differ.  Adhering to the original U-Net publication we used the well studied \emph{softmax} function as our initial output function $p_l^{ \text{softmax} }$ for \emph{cross entropy}:
\begin{equation}
p_l^{ \text{softmax} }(\bm{x}) := \frac{ e^{\mathbf{a}_l(\bm{x}) } }{ \sum\limits_{k \in \mathcal{L}}  e^{ \mathbf{a}_k(\bm{x})}} 
\end{equation}
where $\mathbf{a}_l(\bm{x})$ indicates activation at feature channel $l$ and pixel $\bm{x}\in I$,  i.e., the value that the network takes prior to its final output, therefore here $\mathbf{a}_l(\bm{x}) \in [0,1]$. This definition yields a point estimate for a probability distribution of the $| \mathcal{L} |$-tuple at image position $\bm{x}$ \cite{Goodfellow-et-al-2016}. For the \emph{Dice} contrarily we chose a sigmoid activation function defined as:
\begin{equation}
p_l^{ \text{sigmoid} }(\bm{x}) := \frac{1}{1 + e^{-\mathbf{a}_l(\bm{x})}}
\end{equation}
where $\mathbf{a}_l(\bm{x})$ is defined as above. The final network outputs $p_l^{ \text{softmax} }(\bm{x})$ and $p_l^{ \text{sigmoid} }(\bm{x})$ may be understood as the approximated probabilities of the pixel $\bm{x}$ not belonging to background \cite{Goodfellow-et-al-2016}. Note that in the latter case this is not necessary a probability distribution in the classical sense, and in neither case the class-probabilities need to be unique. Nevertheless, our evaluations indicate that initialization schemes, depth of network, and strong regularization yield neglectable probabilities for ambiguous classes in fully-convolutional architectures when uniqueness is required.  


Building upon the above definition of $p_l^{\text{softmax}}$, the loss function $\Lambda(I, G_I)^{\, \text{cross entropy}}$ may be defined using the distance function
\begin{equation}
d_l ^{ \, \text{cross entropy}}( I, {G_I} ) := \frac{1}{c_{\mathcal{K}}}\sum_{\bm{x} \in I} \chi_{   \pi_l(G_I)} (\bm{x}) \, \log{p_l(\bm{x})}
\end{equation}
where $\chi_{\pi_l(G_I)}(\bm{x})$ is a characteristic function, i.e., $\chi_{\pi_l(G_I)}(\bm{x}) = 1$ iff  $G_I$ is $1$ at position of pixel $\bm{x}$ and $0$ otherwise. This yields what is usually read as the cross entropy between the distributions $ p_l(\bm{x_i})_{ x_i \in I}$ and $ \chi_{\pi_l(G_I)}(\bm{x_i})_{x_i \in I}$.

In conjunction with the above definition of $p_l^{\text{sigmoid}}$, the distance function $d^{\,dice}$ for the Dice coefficient for a training image $I$, a feature channel $l$ and ground-truth mask $G_I$ can then be defined as

\begin{equation}
	d_l ^{\,dice} ( I,   G_I) :=  2 \, \frac{ \sum_{\bm{x} \in I} \chi_{\pi_l(G_I)}(\bm{x}) \, p_l(\bm{x})}{ \sum_{\bm{x} \in I} \left(\chi_{\pi_l(G_I)}(\bm{x}) + p_l(\bm{x})\right)}
\end{equation}

Note that while this definition is analogous to the classic \emph{Dice} overlap measure (compare with \ref{sec:perfMetrics}), working on the real numbers it avoids thresholding at every iteration step and allows for direct implementation in a backpropagation-based optimization algorithm. 

Trying to simplify our set-up we experimented with the chosen activations and their parametrizations. In regards of the activation functions themselves, experiments showed that even relaxed domain restrictions on \emph{cross entropy} as described above, in combination with the chosen loss functions, will still work on a sigmoid output. In order to perform a fair comparison, the following evaluations of the original U-Net \, will stick to the authors choice of \emph{softmax} activations, where in our case we used the \emph{sigmoid} functions.\\






\section{Experimental setup}
\label{sec:exp_setup}

\subsection{Training Details}
\label{sec:dataSet}

We use the JSRT dataset~\cite{Shiraishi2000} both for training and testing. The dataset consists of $247$ posterior-anterior (PA) chest radiographs with a resolution of $2048 \times 2048$, $0.175$ mm pixel size and $12$-bit depth. The reference organ boundaries for JSRT images for left and right lung fields, heart and left and right clavicles were introduced by van Ginneken et al.~\cite{Ginneken2006} in $1024 \times 1024$ resolution and available in the SCR database. 

In order to be able to compare our approach with the state-of-the-art methods, we trained the networks on the two $128 \times 128$ and $256 \times256$ resolutions. Furthermore we derived two slightly distinct ground-truth sets, differing in how they handle the overlapping structures. 
For comparability we built one set of ground-truth masks  $\mathcal{G}_{\text{dice}}$ using the three classes: \emph{left and right lung fields including clavicles}, \emph{left and right clavicles}, and \emph{heart}. This ground-truth set was used for all training runs with the Dice-based loss function. By construction this set-up does not enforce uniqueness of pixel level labelling.  Contrarily, the second ground-truth set $\mathcal{G}_{\text{entropy}}$ was composed using the four classes: \emph{background}, \emph{left and right lung fields without clavicles}, \emph{left and right clavicles}, and \emph{heart}.  $\mathcal{G}_{\text{entropy}}$ was used in all experiments in conjunction with the cross entropy based loss function. Note that while the latter set is an alteration of the original SCR composition, it was used in training runs only within the study. 
In every evaluation the dataset was split into the same subsets $\mathbf{I}_{\, \text{TRAIN}}$, $\mathbf{I}_{\, \text{VALID}}$ and $\mathbf{I}_{\, \text{TEST}}$. At the testing the same set $\mathbf{I}_{\, \text{TEST}}$ was used for computing performance scores for networks trained with both loss functions.

All images were zero-centered first by subtracting the mean and then additionally normalized by scaling using the standard deviation. The mean and standard deviation were computed across the whole training dataset.

We trained the networks using an ADAM \cite{Kingma2014} optimization algorithm with a fixed initial rate of $10^{-5}$ and the standard values of $\beta_1=0.9$ and $\beta_2=0.999$. We also performed experiments with larger learning rates however training runs were less stable and often lead to local minima or trivial solutions.


\subsection{Performance Metrics}
\label{sec:perfMetrics}

\begin{table*}[ht!]
	\centering
	\renewcommand{\arraystretch}{1.3}
	\begin{minipage}{.48\textwidth}
		\begin{tabular}{|c|c|c|c|c|c|c|}
			\hline
			Body Part                                     & \multicolumn{2}{c|}{Lungs}          & \multicolumn{2}{c|}{Clavicles}      & \multicolumn{2}{c|}{Heart}          \\ \hline \hline
			Evaluation Metric                            & $D$              & $J$              & $D$              & $J$              & $D$              & $J$              \\ \hline \hline
			\invertednet                                 & 0.972            & 0.946            & \textbf{0.902} & \textbf{0.821} & 0.935            & 0.879            \\ \hline
			\udropall                                    & \textbf{0.973} & \textbf{0.948} & 0.896            & 0.812            & \textbf{0.941} & \textbf{0.888} \\ \hline
			\uallconv                                    & 0.971            & 0.944            & 0.876            & 0.780            & 0.938            & 0.883            \\ \hline
			Original U-Net                               & 0.971            & 0.944            & 0.880            & 0.785            & 0.938            & 0.883            \\ \hline
		\end{tabular}
		\caption{Evaluation results of four compared architectures. All scores are computed on the testing set in a three-fold cross-validation manner with networks trained with the loss function based on the cross entropy distance at $256\times256$ imaging resolution.}
		\label{table:evaluations_256}
	\end{minipage}
	\hfill
	\begin{minipage}{.48\textwidth}
		\begin{tabular}{|c|c|c|c|c|c|c|}
			\hline
			Body Part                                     & \multicolumn{2}{c|}{Lungs}          & \multicolumn{2}{c|}{Clavicles}      & \multicolumn{2}{c|}{Heart}          \\ \hline \hline
			Evaluation Metric & $D$              & $J$              & $D$              & $J$              & $D$              & $J$              \\ \hline \hline
			\invertednet      & \textbf{0.966} & \textbf{0.934} & \textbf{0.889} & \textbf{0.801} & \textbf{0.940} & \textbf{0.888} \\ \hline
			\udropall         & 0.965            & 0.932            & 0.837            & 0.720            & 0.929            & 0.868            \\ \hline
			\uallconv         & 0.965            & 0.932            & 0.834            & 0.715            & 0.928            & 0.866            \\ \hline
			Original U-Net    & 0.964            & 0.930            & 0.834            & 0.716            & 0.934            & 0.877            \\ \hline
			
		\end{tabular}
		\caption{Evaluation results of four compared architectures. All scores are computed on the testing set in a three-fold cross-validation manner with networks trained with the loss function based on the cross entropy distance at $128\times128$ imaging resolution.}
		\label{table:evaluations_128}
	\end{minipage}\hfill
	
\end{table*}

To evaluate the architectures and compare with state-of-the-art approaches, we used the Dice ($D$) and Jaccard ($J$) similarity coefficients, defined as follows. 

Given an image $I$, ground-truth mask $G_I$ and the feature channel $l$, let $P_l(I)$ be the set of pixels where the model is certain that they do not belong to the background, i.e.,
\begin{equation}
P_l(I):=\left\{\bm{x}: \bm{x} \in I \, \land \, | \, p_l(\bm{x}) - 1 \, | < \epsilon \right\}
\label{eq:pixel_set}
\end{equation}
where $\epsilon=0.25$ is an empirically chosen threshold value.

The similarity coefficients $D$ and $J$ might then be computed in the following way:
\begin{equation}
D ( I,  G_I) := 2 \, \frac{| \, P_l(I) \cap  \pi_l(G_I)  \, |}{|P_l(I)| + |\, \pi_l(G_I)|}
\end{equation}
\begin{equation}
J( I,  G_I) = \frac{2}{2-D( I,  G_I)}
\end{equation}

In addition, for our best architecture we computed the symmetric mean absolute surface distance $(S_d)$ as defined by Babalola et al.~\cite{Babalola2009}. 


\subsection{Implementation Details}

All experiments were performed using Keras \cite{Chollet2015} with Theano backend \cite{Theano2016} in Python 3.5. The backend was used for automatic differentiation and optimization during training.

We used zero-padding in convolutional layers in all architectures. Therefore, output channels have the same dimensions as the input. To reduce the number of parameters and speed up training, instead of the last dense layer in each network we used a convolutional layer, with the number of feature maps equal to the number of considered classes for the loss function based on the Dice coefficient and with one more for background for the loss function based on the cross entropy. $S_d$ distances reported in the evaluations were computed using MedPy Python library.

Downsampling of the original masks of the SCR database was performed using the \emph{scikit-image} Python library \cite{scikit-image}

\section{Results and Discussion}

\label{sec:results}

\subsection{Multi-class segmentation with loss function based on cross-entropy}

Table \ref{table:evaluations_256} contains a comparison between U-Net and our three proposed architectures when trained with a loss function based on cross entropy in a three-fold cross-validation scheme. We chose this function for this part of our experiments to compare the performance of the proposed architectures with the U-Net.

Scores for \textit{lung segmentation} did not vary significantly across the architectures. \udropall \, and \invertednet \, slightly outperformed the U-Net. \textit{Clavicle segmentation} was a more challenging task for all our architectures. This is not surprising, as clavicles are much smaller than hearts and lungs and their shapes vary more. \udropall \, and \invertednet \, outperformed the U-Net by 2.7\% and 3.6\% in Jaccard overlap score. On the \textit{heart segmentation} task, the results did not vary significantly, though the \udropall \, slightly outperformed other architectures.

\begin{table*}[ht!]
	\centering
	\renewcommand{\arraystretch}{1.3}
	\begin{tabular}{|c|c|c|c|c|c|c|}
		\hline
		Body Part                                     & \multicolumn{2}{c|}{Lungs}          & \multicolumn{2}{c|}{Clavicles}      & \multicolumn{2}{c|}{Heart}          \\ \hline \hline
		Evaluation Metric                            & $D$              & $J$              & $D$              & $J$              & $D$              & $J$              \\ \hline \hline
		\invertednet \, + \uallconv \, + \udropall         & 0.973                & 0.948            & 0.902                & 0.822            & 0.940                & 0.887            \\ \hline 
		\invertednet \, + \uallconv            &          0.973       & 0.948            & 0.901                & 0.820            & 0.938                 & 0.883            \\ \hline
		\invertednet \, + \udropall         & \textbf{0.974}                &\textbf{0.949}            & \textbf{0.910}                & \textbf{0.833}            & \textbf{0.941}                & \textbf{0.888} \\ \hline
		\uallconv \, + \udropall & 0.972                & 0.946 &   0.892              & 0.805                & 0.935                & 0.878                \\ \hline
		  
	\end{tabular}
	\caption{Evaluation results of ensembles of networks on the combination of the three proposed architectures at $256\times256$ imaging resolution.}
	\label{table:evaluations_ensembles}
\end{table*}

On the $128\times128$ imaging resolution (scores shown in Table~\ref{table:evaluations_128}), the \invertednet \, displayed the best performance on the clavicle segmentation where the original U-Net was surpassed by 8.5\% in Jaccard overlap score. In summary, clavicles seem to be more challenging for the original U-Net and its more similar modifications \uallconv \, and \udropall \, at the lower resolution because of the higher number of pooling layers and smaller number of low level abstract features in the contraction part of the network architecture.

\begin{figure}[h!]
	\includegraphics[width=0.48\textwidth]{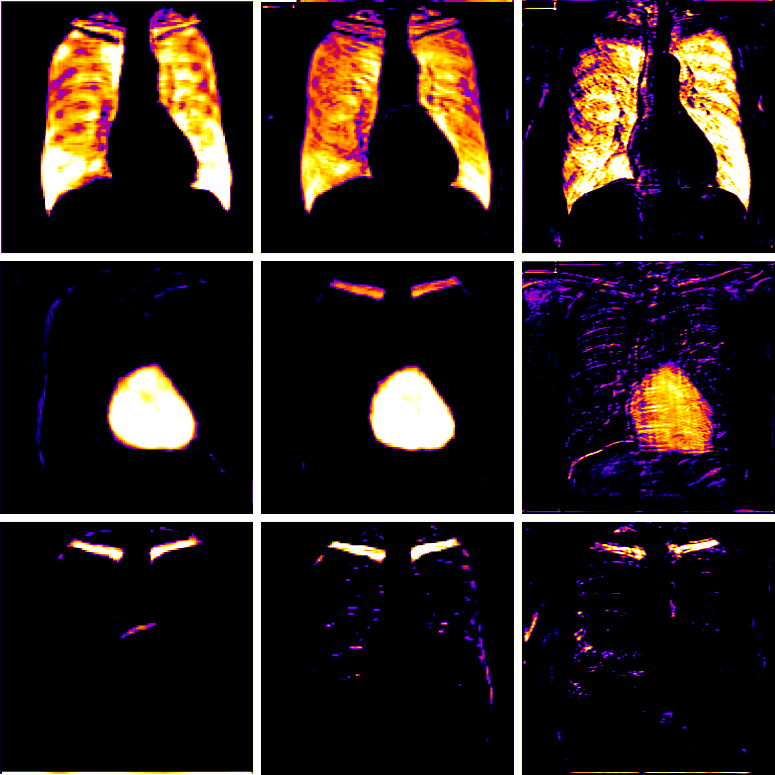}
	\caption{Examples of features extracted after the penultimate upsampling step for \udropall \, (left), \uallconv \, (center) and \invertednet \, (right). The same test image was use in all three cases. Higher colour intensities correspond to higher activation values.}
	\label{fig:fMaps}
\end{figure}

Visualization of the derived features of our proposed networks (shown in Fig.~\ref{fig:fMaps}) shows that compared to \udropall \, (left) and \uallconv \, (center), the \invertednet \, (right) favours sharper borders over a good shape separation. In combination with the higher feature map count just before the classification step this yields a higher overlap score. Nevertheless the different areas of network \emph{specialization} indicate that an ensemble of the networks might improve the results further. All images were derived from the same test image after the penultimate upsampling step of the networks.
		
Motivated by the successful implementation of ensemble networks \cite{Ju2017, Hu2017}, and described visual inspection of the derived features of our networks, we additionally evaluated ensembles of pairs and the triplet consisting of the proposed architectures. Final scores are shown in Table~\ref{table:evaluations_ensembles}. The driving idea is to compose the end result out of the emphasized strong parts of the different architectures. Resulting masks for the ensemble networks were computed via averaging the outputs of the evaluated networks and further thresholding the result using majority voting.
The pair of the two best performing architectures \invertednet \, and \udropall \, slightly improves the scores on all organs with the largest difference achieved on the clavicles.

\begin{figure*}[th!]
	\centering
	\begin{subfigure}{0.31\textwidth}
		\includegraphics[width=\textwidth]{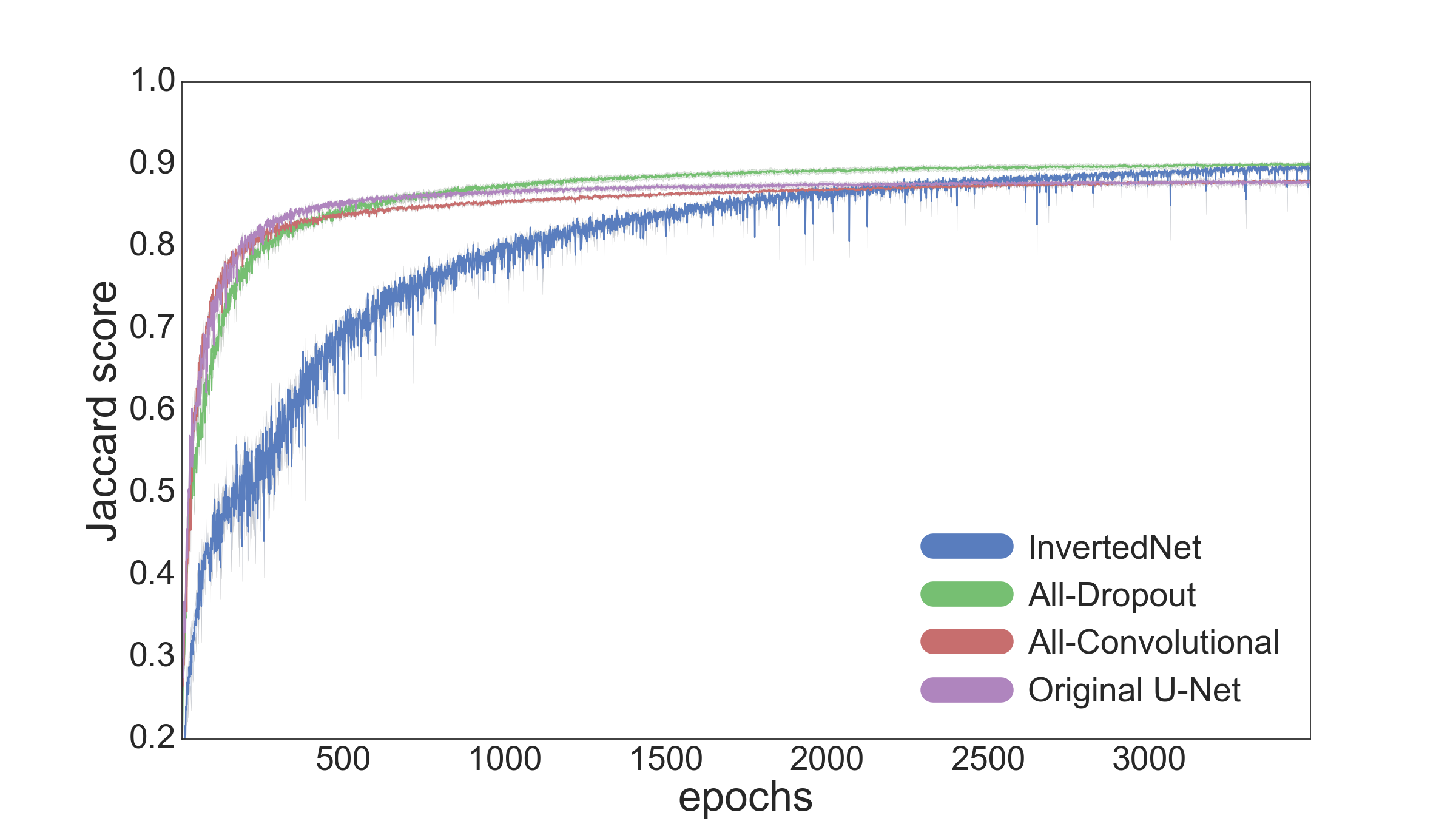}
		\label{fig:clavicles_convergence_0_256}
		\caption{}
	\end{subfigure}
	\begin{subfigure}{0.31\textwidth}
		\includegraphics[width=\textwidth]{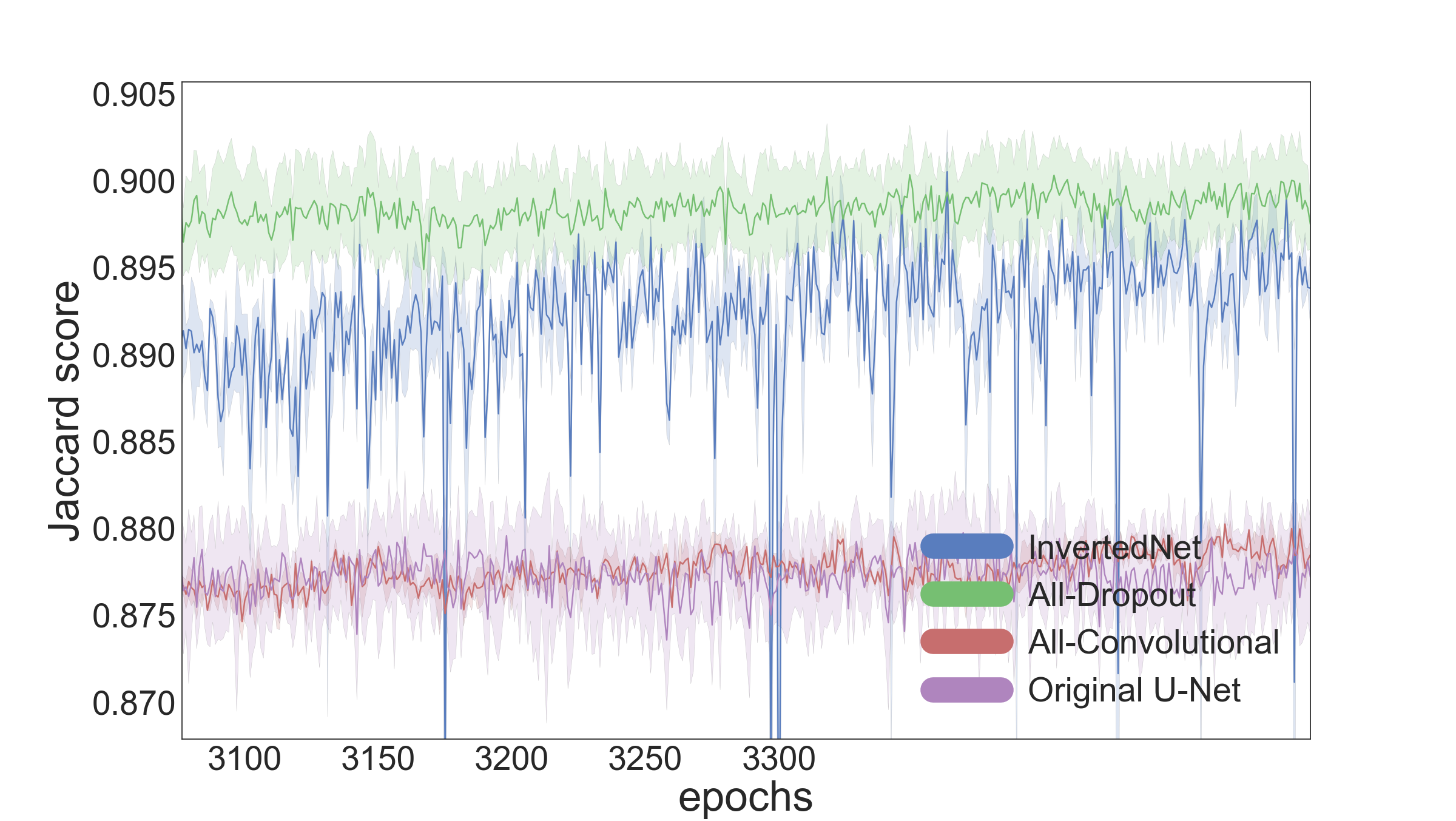}
		\label{fig:clavicles_convergence_last_256}
		\caption{}
	\end{subfigure}
	\begin{subfigure}{0.31\textwidth}
		\includegraphics[width=\textwidth]{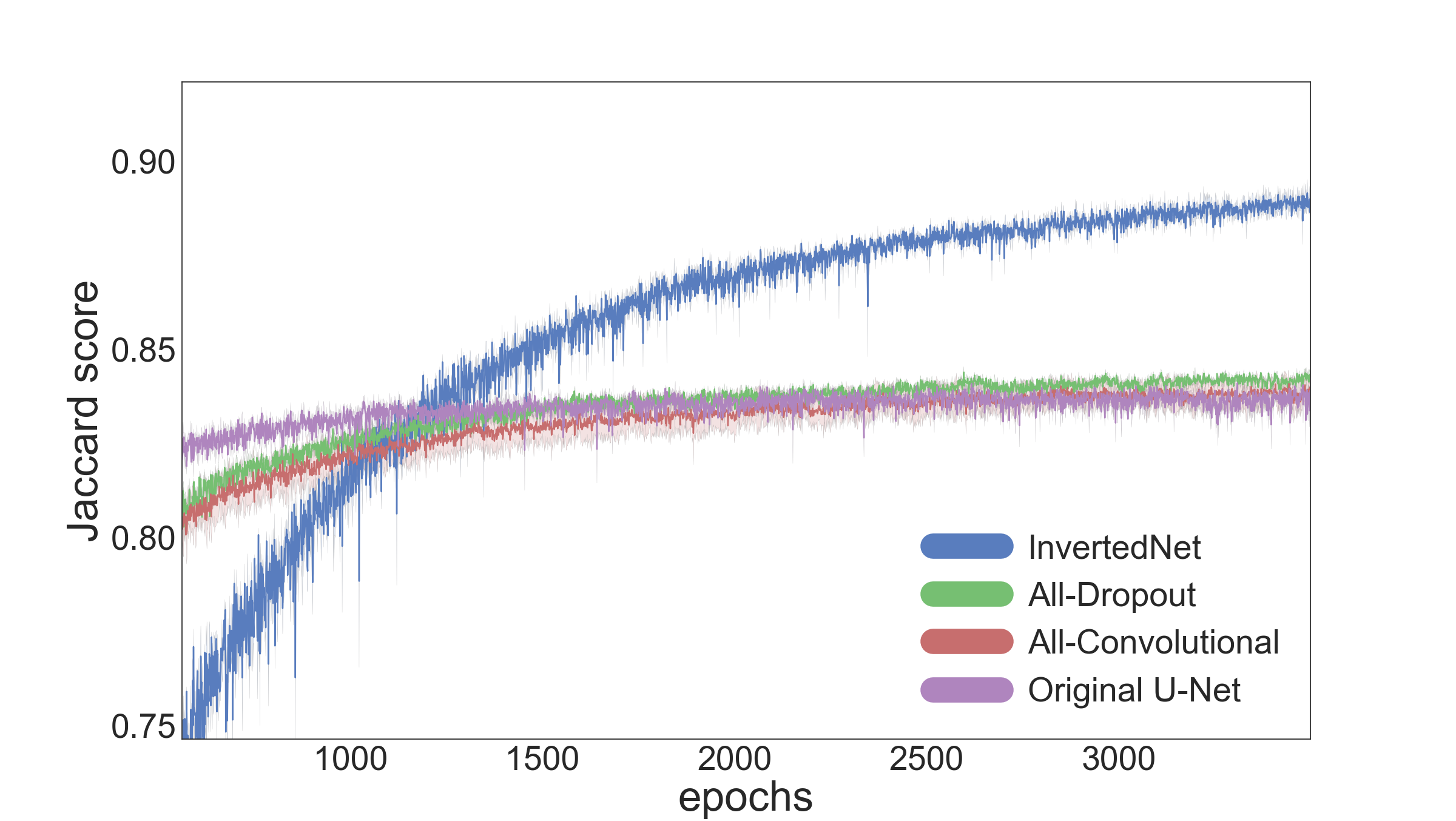}
		\label{fig:clavicles_convergence_last_128}
		\caption{}
	\end{subfigure}

	\caption{Performance of models on the validation set during training for clavicles: in the initial (a) and final (b) epochs of the training for $256\times256$ resolution and c) in the final epochs of the training for $128\times128$ resolution}
	\label{fig:clavicles_convergence}
\end{figure*}
\begin{figure*}[ht!]
	\centering
	\includegraphics[width=0.6\textwidth]{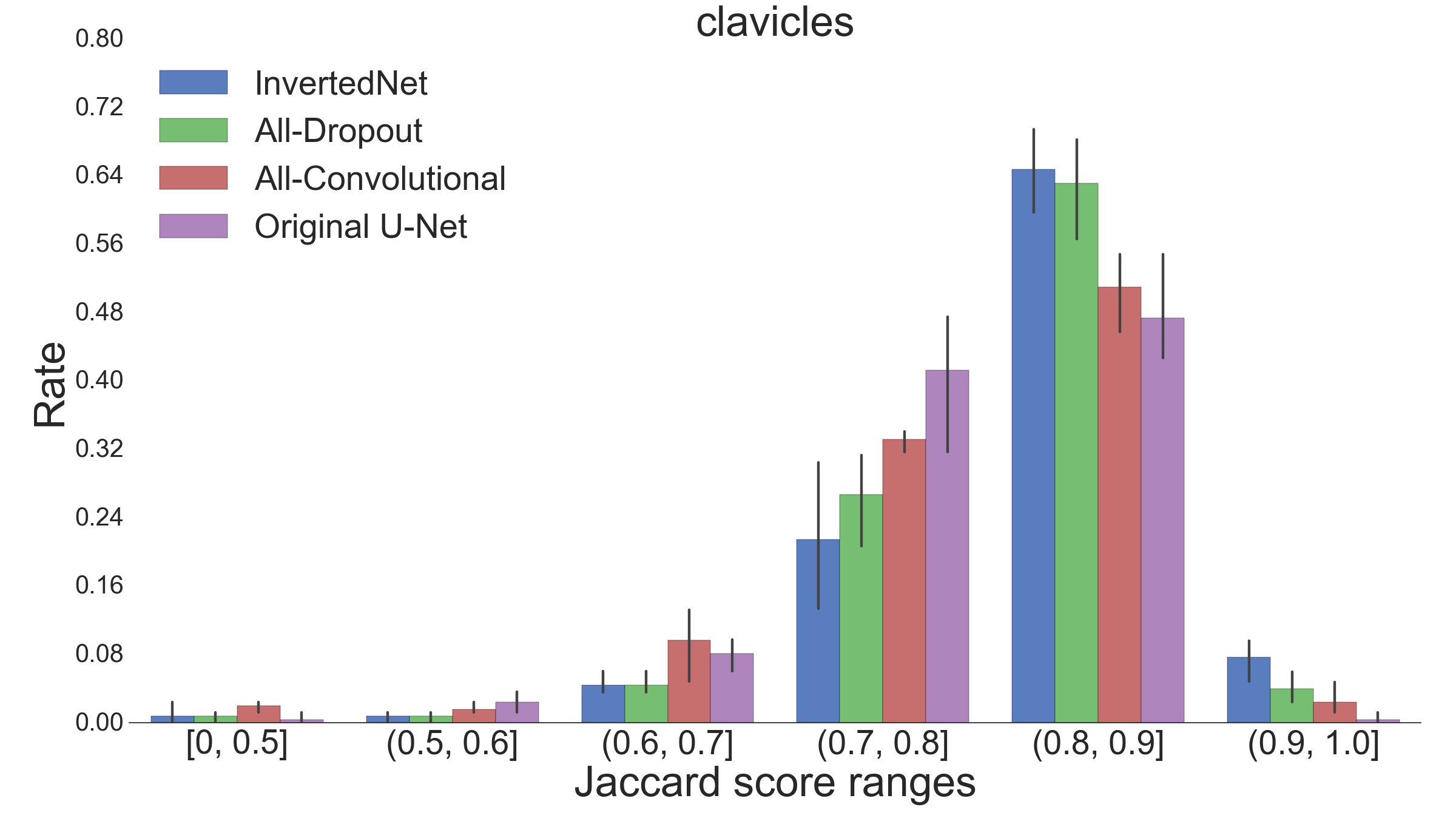}
	\caption{The percentage contribution (\textit{y}-axis) of each model (columns) to each range of Jaccard score (\textit{x}-axis) on the test set}
	\label{fig:clavicles_ranges}
\end{figure*}
\begin{table*}[ht!]
	\centering
	\renewcommand{\arraystretch}{1.2}
	\begin{tabular}{|c|c|c|c|c|c|c|c|c|c|c|c|c|}
		\hline
		Loss Function                  & \multicolumn{6}{c|}{Weighted Loss Function with Dice}                                         & \multicolumn{6}{c|}{Non-Weighted Loss Function with Dice}                                     \\ \hline
		Body Part                       & \multicolumn{2}{c|}{Lungs} & \multicolumn{2}{c|}{Clavicles} & \multicolumn{2}{c|}{Heart} & \multicolumn{2}{c|}{Lungs} & \multicolumn{2}{c|}{Clavicles} & \multicolumn{2}{c|}{Heart} \\ \hline
		Metric                         & $D$          & $J$             & $D$            & $J$               & $D$          & $J$             & $D$          & $J$             & $D$            & $J$               & $D$          & $J$             \\ \hline \hline
		Training 60\%, Validation 7\%  &     0.972       & 0.946         &      0.930        & 0.870           &     0.931       & 0.871         &   0.975         & 0.951         &       0.918       & 0.848           &     0.925       & 0.861         \\ \hline
		Training 50\%, Validation 17\% &    0.971         & 0.945         &      0.922        & 0.856           &     0.927       & 0.865         &      0.974      & 0.949         &       0.915       & 0.844           &     0.932       & 0.874         \\ \hline
		Training 45\%, Validation 22\% &      0.970      & 0.942         &    0.923           & 0.858           &     0.932       & 0.874         &    0.974        & 0.949         &       0.916       & 0.846           &     0.923       & 0.857         \\ \hline
	\end{tabular}
	\caption{Evaluation comparison of InvertedNet architecture for different training and validation splits for the weighted and non-weighted loss functions based on the Dice coefficient. In all three evaluations, the same set containing 33\% of images from JSRT dataset was used for testing}
	\label{table:validationsplits_dice_weighted_non_weighted}
\end{table*}
\begin{table*}[h!]
	\centering
	\renewcommand{\arraystretch}{1.2}
\begin{tabular}{|c|c|c|l|c|c|l|c|c|c|}
	\hline
	Body Part                      & \multicolumn{3}{c|}{Lungs} & \multicolumn{3}{c|}{Clavicles} & \multicolumn{3}{c|}{Heart} \\ \hline \hline
	Evaluation Metric              & $D$       & $J$       & $S_{d}$      & $D$         & $J$        & $S_{d}$       & $D$       & $J$       & $S_{d}$      \\ \hline
	Training 60\%, Validation 7\%  & 0.975   & 0.952   & 0.67   & 0.931     & 0.872    & 0.72    & 0.931   & 0.871   & 1.98   \\ \hline
	Training 50\%, Validation 17\% & 0.974   & 0.950   & 0.69   & 0.929     & 0.868    & 1.38    & 0.937   & 0.882   & 1.94   \\ \hline
	Training 45\%, Validation 22\% & 0.973   & 0.948   & 0.71   & 0.924     & 0.859    & 2.10    & 0.935   & 0.878   & 2.01   \\ \hline
\end{tabular}
	\caption{Evaluation comparison of InvertedNet architecture with ELU activation functions for different training and validation splits for the loss function based on the Dice coefficient. In all three evaluations, the remaining 33\% of images are used for testing.}
	\label{table:validationsplits_dice_elu_relu}
\end{table*}

\begin{table*}[ht!]
	\centering
	\renewcommand{\arraystretch}{1.3}
	\begin{tabular}{|c|c|c|c|c|c|c|}
		\hline
		Body Part                                     & \multicolumn{2}{c|}{Lungs}          & \multicolumn{2}{c|}{Clavicles}      & \multicolumn{2}{c|}{Heart}          \\ \hline \hline
		Evaluation Metric                            & $D$              & $J$              & $D$              & $J$              & $D$              & $J$              \\ \hline \hline
		Human Observer~\cite{Ginneken2006}         & -                & 0.946            & -                & 0.896            & -                & 0.878            \\ \hline \hline
		ASM Tuned~\cite{Ginneken2006}  (*)            & -                & 0.927            & -                & 0.734            & -                & 0.814            \\ \hline
		Hybrid Voting~\cite{Ginneken2006} (*)          & -                & 0.949            & -                & 0.736            & -                & 0.860 \\ \hline
		Ibragimov et al.~\cite{Ibragimov2016} & -                & \textbf{0.953} & -                & -                & -                & -                \\ \hline
		Seghers et al.~\cite{Seghers2007}     & -                & 0.951            & -                & -                & -                & -                \\ \hline
		\invertednet \, with ELU   & 0.974            & 0.950            & 0.929 & \textbf{0.868} & 0.937            & \textbf{0.882}            \\ \hline
	\end{tabular}
	\caption{Our best architecture compared with state-of-the-art methods; (*) single-class algorithms trained and evaluated for different organs separately; "-" the score was not reported}
	\label{table:evaluations_comparison}
\end{table*}


Fig.~\ref{fig:clavicles_convergence} shows how the proposed models and the U-Net perform on the validation set at each epoch during training. All models were trained in a three-fold manner. The average validation curves are shown in bold and the shading corresponds to the standard deviations.  
The three figures depict convergence plots for two resolutions ($128\times128$ and $256\times256$). The overlap scores of the U-Net typically grow faster than the other networks at the start, but then plateau and oscillate until the end of the training procedure. Other, better regularized architectures start off slower (depicted at Fig.~\ref{fig:clavicles_convergence}a), but ultimately reach similar or higher scores (shown in Figs.~\ref{fig:clavicles_convergence}b and \ref{fig:clavicles_convergence}c). \invertednet \, starts extremely slowly, but ultimately achieves the best result.  

Fig.~\ref{fig:clavicles_ranges} shows performance results for U-Net and the three proposed architectures on the clavicle segmentation task. The \textit{x}-axis corresponds to binned intervals for Jaccard overlap scores, while the \textit{y}-axis corresponds to percentages of test samples falling into the Jaccard intervals from the \textit{x}-axis. The factor plot was produced using overlap scores achieved in a three-fold cross-validation manner on the testing set. \invertednet \, has the most samples in the last range and the fewest number of samples falling into ranges $<$ 0.6. Jaccard scores for more than half of the testing samples for \invertednet \, yielded values greater than 0.8. 

In order to support our choice of the best performing architecture we additionally analyse how the segmentation results of the compared models differ in terms of statistical significance test scores. Table~\ref{table:significance} shows the segmentation result significance analysis using Wilcoxon signed-rank test for Jaccard scores on the test set. The p-values are given for lungs, clavicles and heart. The entries of the table with values $<0.01$ correspond to pairs of architectures demonstrating statistically significant difference in segmentation results on the test set. Therefore, the shown p-values complement and confirm the results shown in Table~\ref{table:evaluations_256} and Table~\ref{table:evaluations_comparison}. Statistical significance tests show that comparing to the other architectures the difference is significant for lungs and clavicles in case of \invertednet , and lungs, clavicles and heart in case of \udropall. However, based on both evaluations, we choose the \invertednet \, as our winning model with which we perform the following additional evaluations. 

\begin{figure*}[t!]
	\centering
	\begin{subfigure}{0.19\textwidth}
		\includegraphics[width=\textwidth]{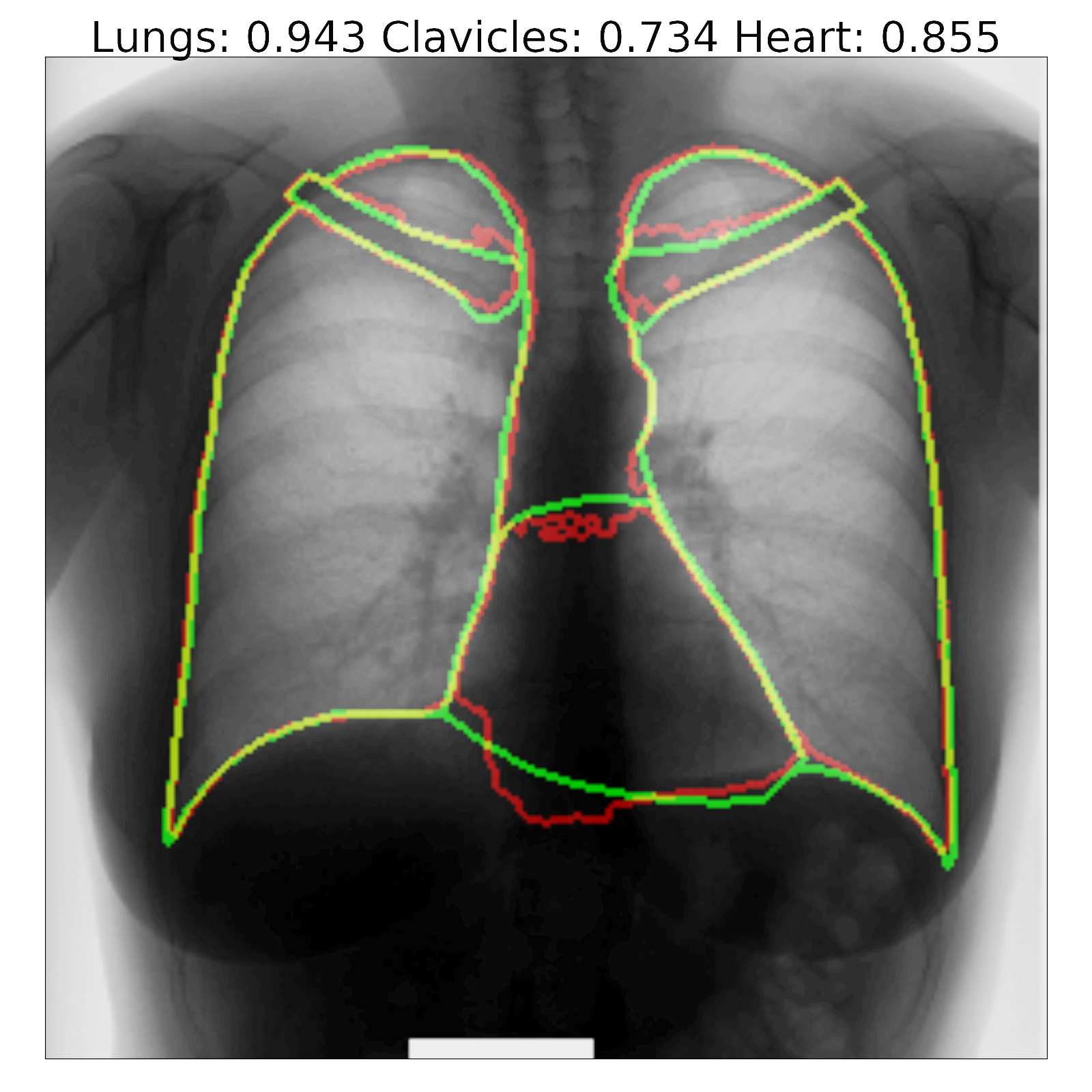}
	\end{subfigure}
	\begin{subfigure}{0.19\textwidth}
		\includegraphics[width=\textwidth]{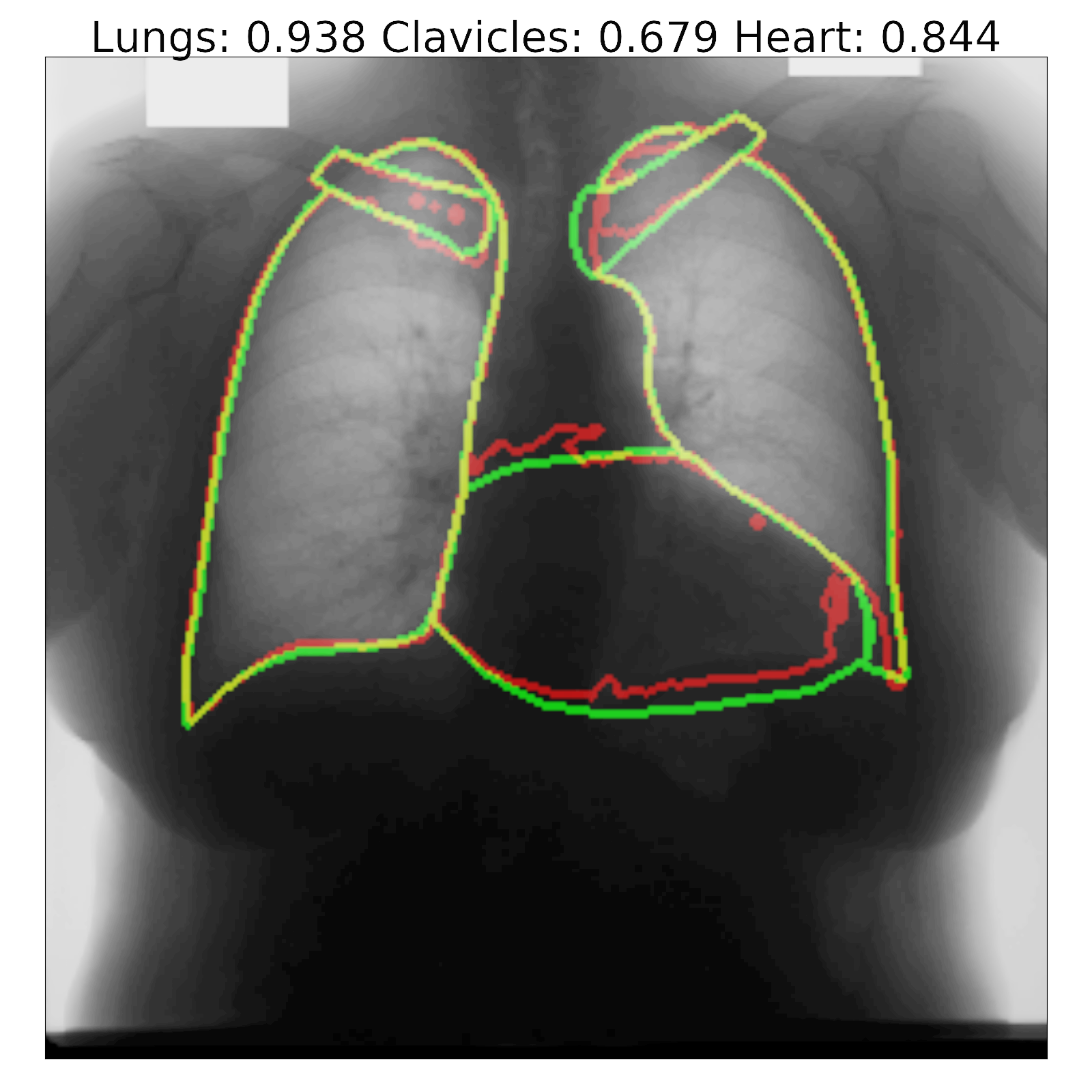}
	\end{subfigure}
	\begin{subfigure}{0.19\textwidth}
		\includegraphics[width=\textwidth]{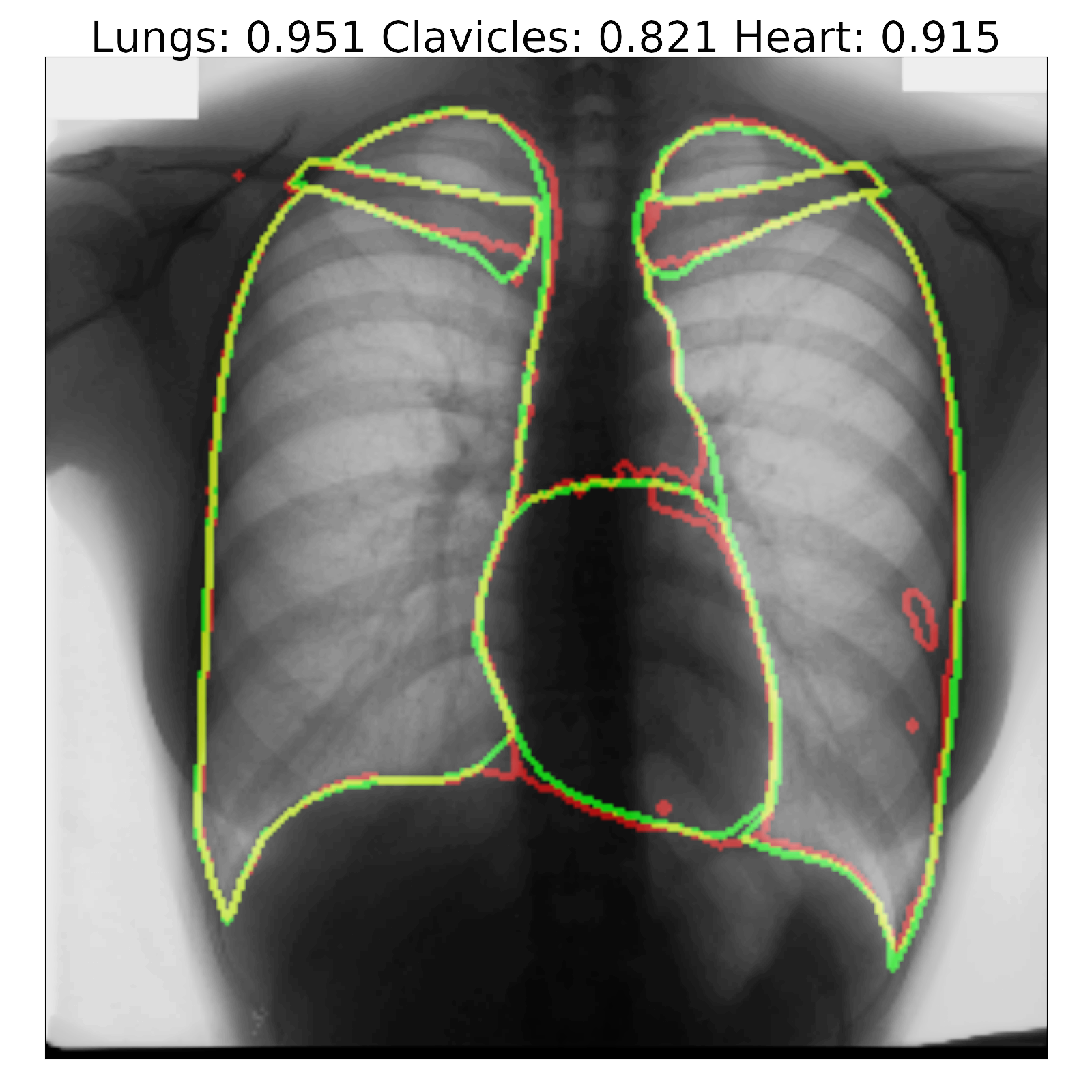}
	\end{subfigure}
	\begin{subfigure}{0.19\textwidth}
		\includegraphics[width=\textwidth]{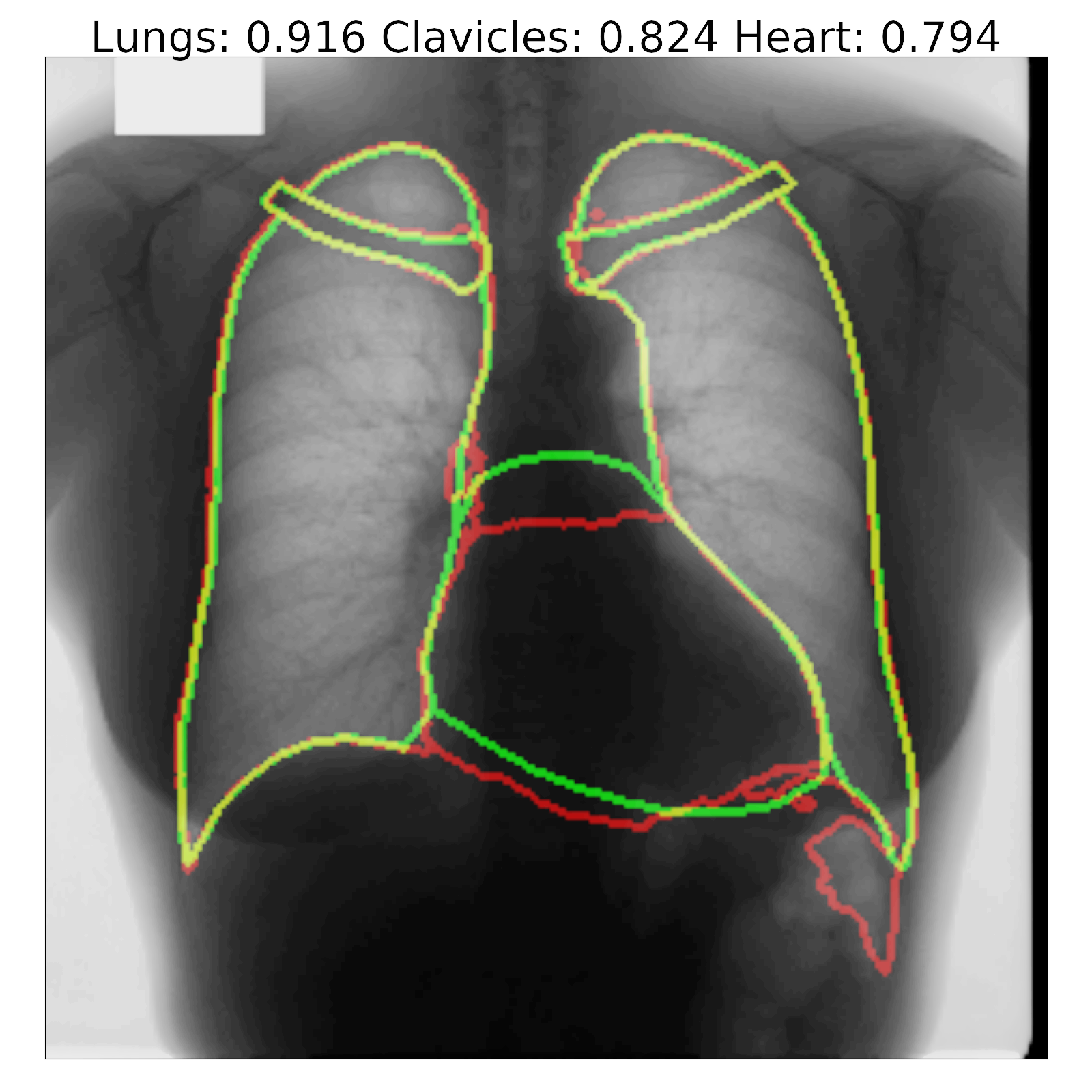}
	\end{subfigure}
	\begin{subfigure}{0.19\textwidth}
		\includegraphics[width=\textwidth]{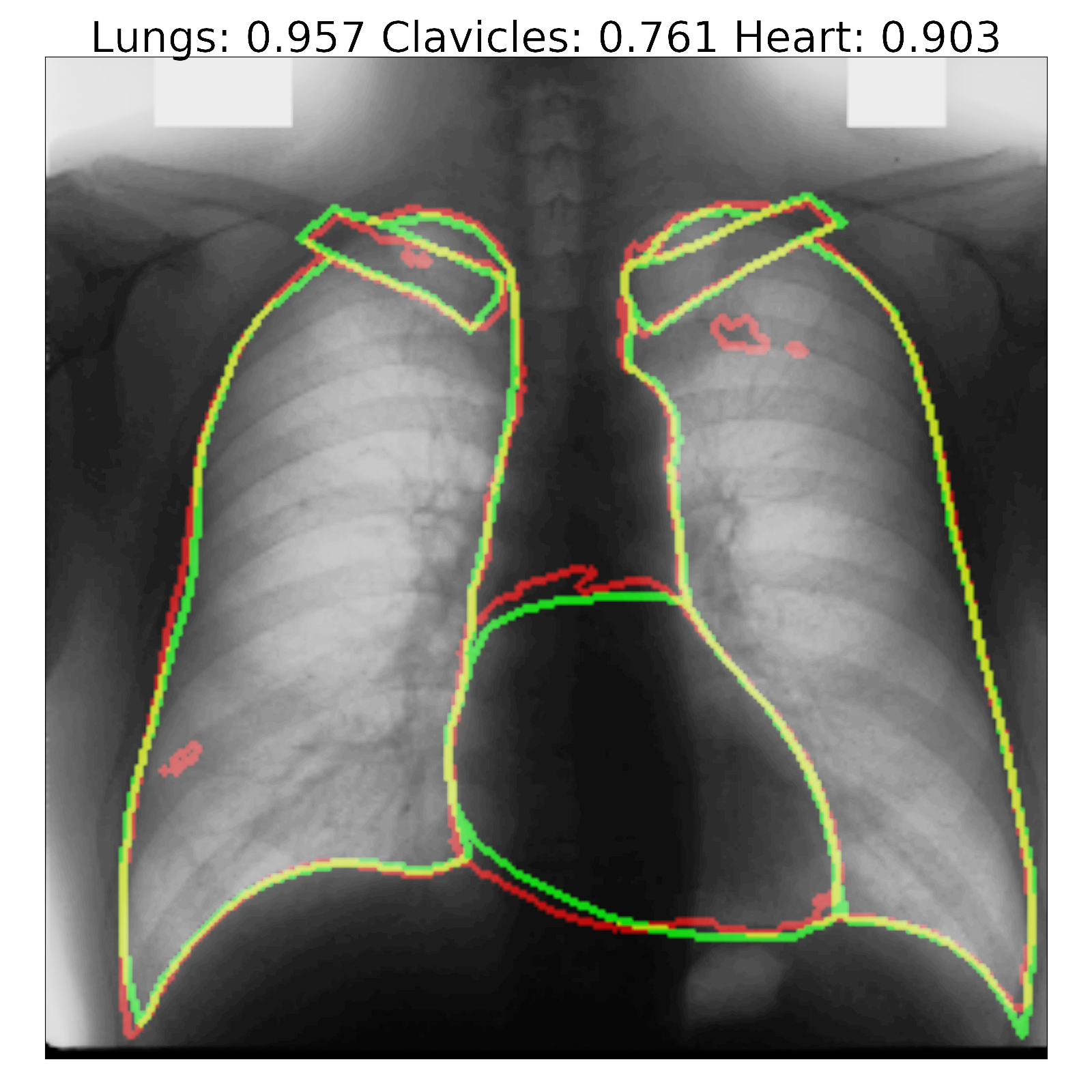}
	\end{subfigure}
	
	\begin{subfigure}{0.19\textwidth}
		\includegraphics[width=\textwidth]{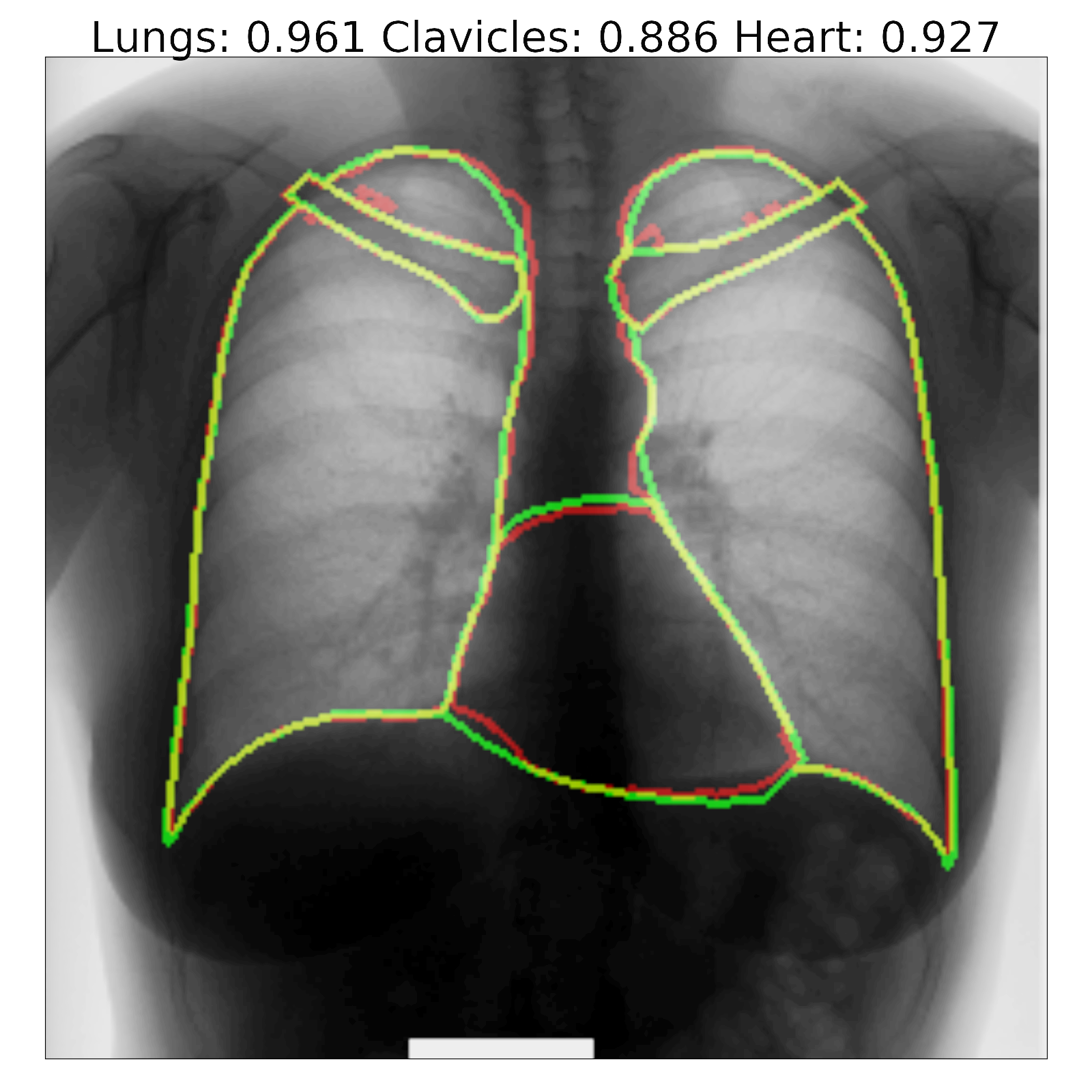}
	\end{subfigure}
	\begin{subfigure}{0.19\textwidth}
		\includegraphics[width=\textwidth]{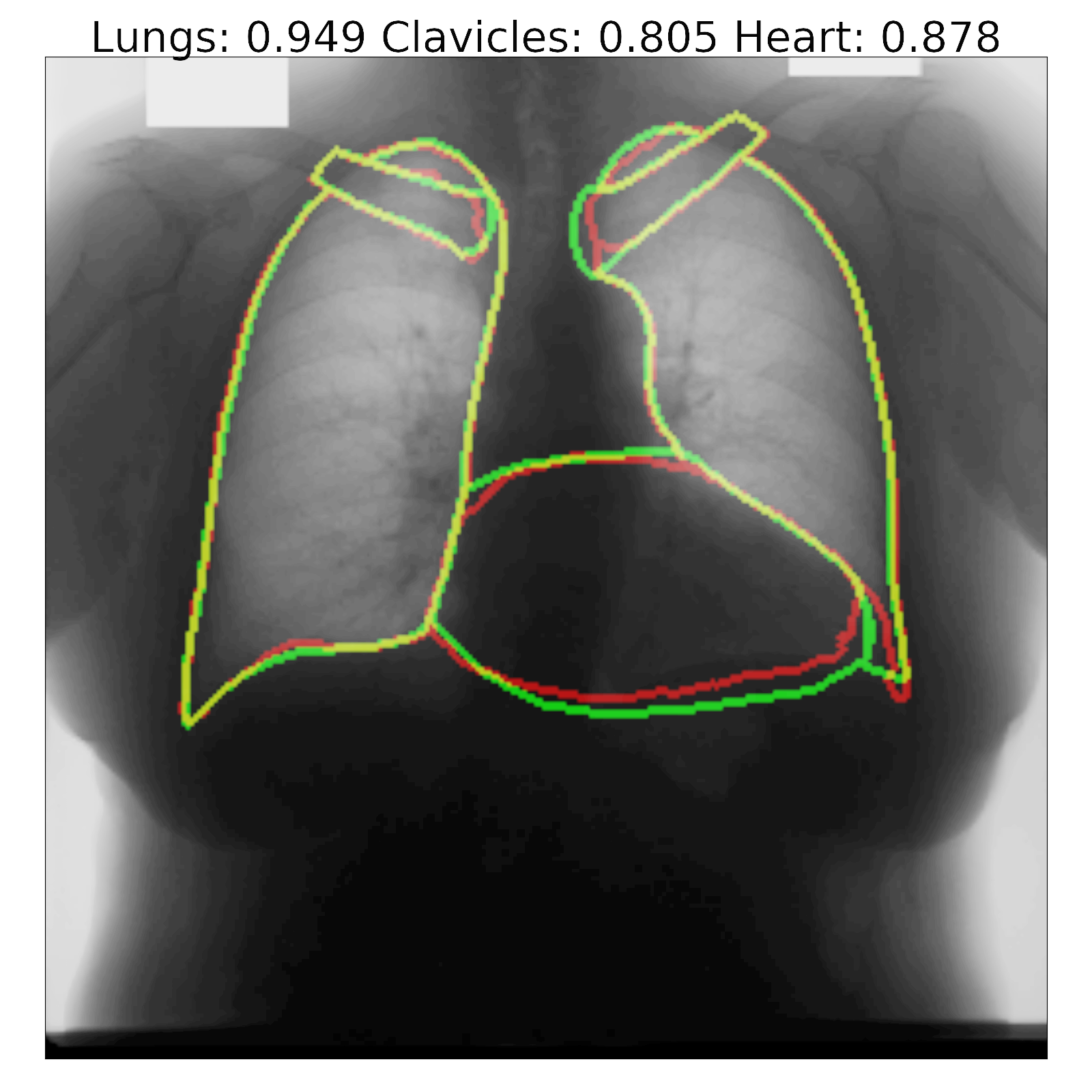}
	\end{subfigure}
	\begin{subfigure}{0.19\textwidth}
		\includegraphics[width=\textwidth]{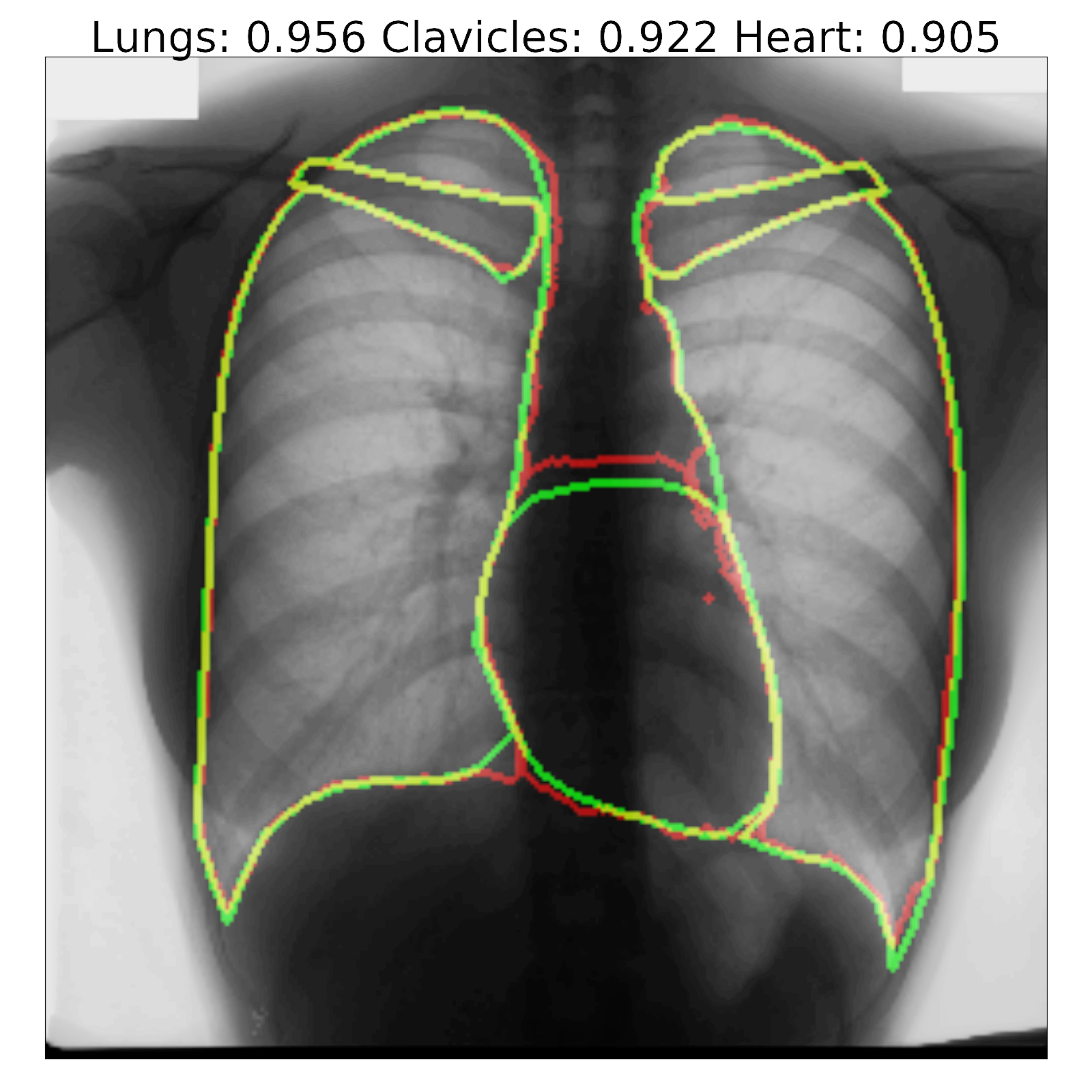}
	\end{subfigure}
	\begin{subfigure}{0.19\textwidth}
		\includegraphics[width=\textwidth]{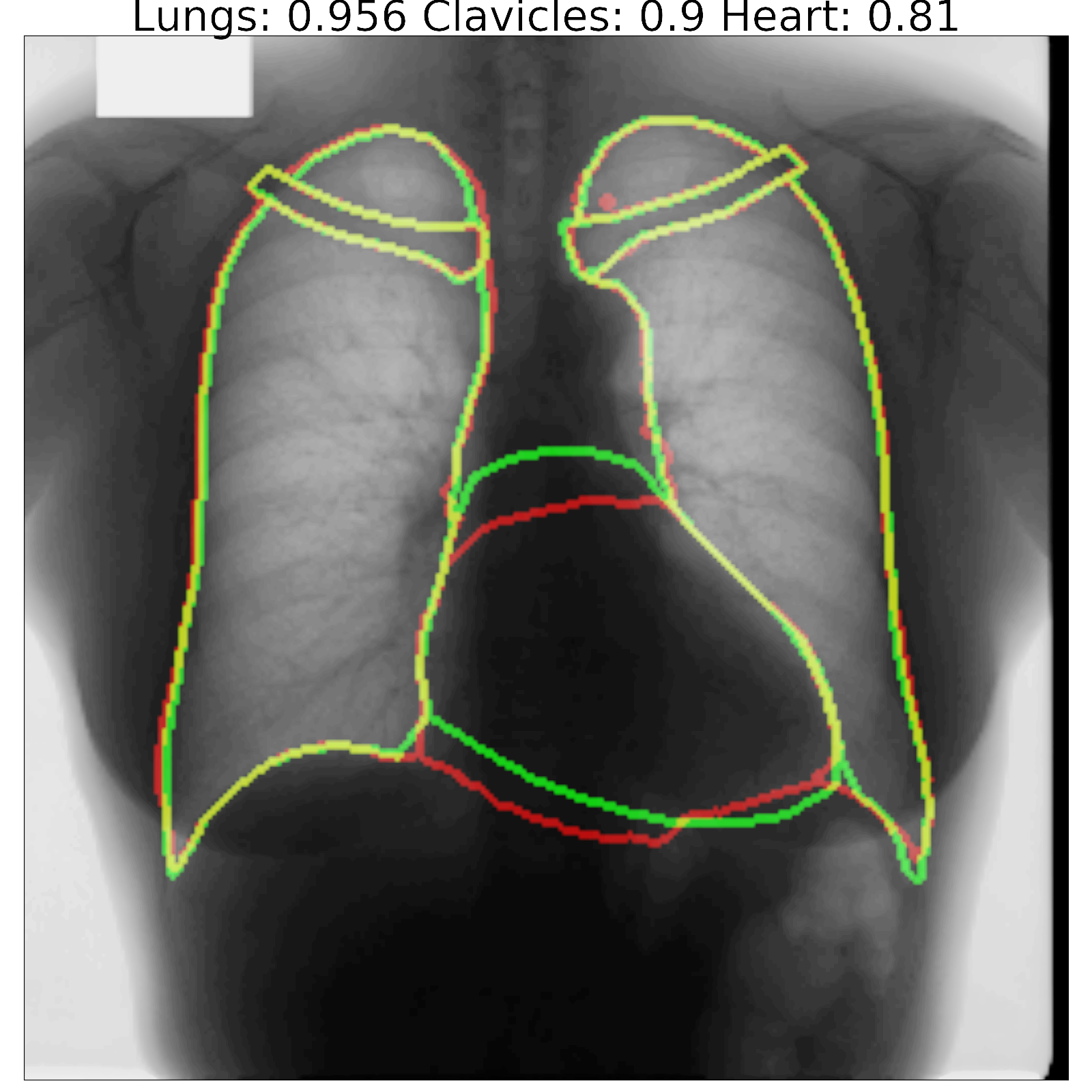}
	\end{subfigure}
	\begin{subfigure}{0.19\textwidth}
		\includegraphics[width=\textwidth]{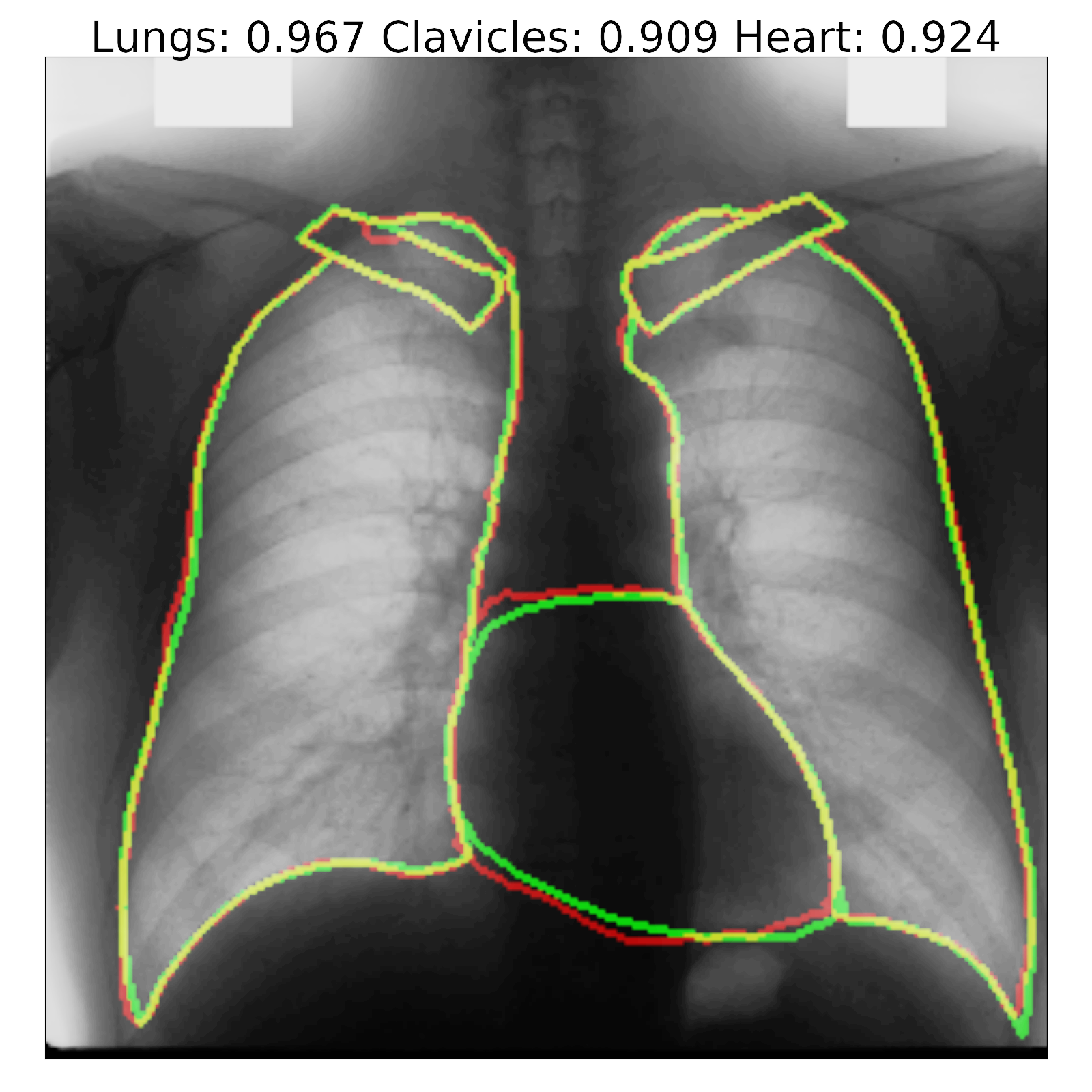}
	\end{subfigure}
	
	\caption{Segmentation results and corresponding Jaccard scores on some images for U-Net (top row) and proposed InvertedNet with ELUs (bottom row). The contour of the ground-truth is shown in green, segmentation result of the algorithm in red and overlap of two contours in yellow.}
	\label{figure:testing_segmentations}
\end{figure*}

\subsection{Multi-class segmentation with loss function based on Dice coefficient}

The performance of the overall best architecture \invertednet, has been evaluated with several splits of input data into training and validation sets and for two loss functions (with and without class weights) based on the Dice coefficient. The same testing set was used in all training runs. The results are shown in Table~\ref{table:validationsplits_dice_weighted_non_weighted}. Despite the assumption that the loss function based on the Dice coefficient does not require any weighting, we show in this evaluation that this is not always the case. In the presence of severe between-class imbalance in the data, it is still important to use class weighting. However, the dominant class slightly benefits from non-weighting of the loss function. 
Nevertheless, this table shows that in contrast to the cross entropy coefficient, using the Dice coefficient for the loss function can improve the final score for clavicles by more than 4\%. 
\subsection{ReLU vs ELU activation functions}

Motivated by the work by Clevert et al.~\cite{Clevert2015}, we also evaluated performance of a modification of the \invertednet \, with exponential linear units (ELUs). We ran evaluations with three different training / validation test splits as in the case with rectified linear units (ReLUs). Table~\ref{table:validationsplits_dice_elu_relu} shows scores for this evaluation. Using ELUs resulted in a clear increase in overlap scores for all organs, with the most significant improvements achieved for clavicles and heart. 

Fig.~\ref{figure:testing_segmentations} shows a few examples of the algorithm results for both successful and failed cases for U-Net (top) and the \invertednet \, trained with the ELU and loss function based on the Dice coefficient. The white boxes show Jaccard scores for lungs, clavicles and heart. To extract the shape contours of the segmentation and ground-truth, we used a morphological outline extraction algorithm on both segmentation result and reference masks. The contour of the ground-truth is shown in green, the segmentation result of the algorithm in red and the overlap of two contours in yellow colors.

\subsection{Comparison with state-of-the-art methods}

Table~\ref{table:evaluations_comparison} shows the comparison between the Human observer, state-of-the-art methods and \invertednet \, with ELU activation functions when trained with the loss function based on the Dice coefficient. While \invertednet \, could not surpass the best approaches by Ibragimov et al.~\cite{Ibragimov2016} and Seghers et al.~\cite{Seghers2007}, it outperformed the human observer on the lung segmentation task. \invertednet \, outperformed the best state-of-the-art method by 2.2\% in Jaccard overlap score and even slightly beat the human observer on the heart segmentation task. Though \invertednet \, could not outperform human performance on the clavicle segmentation, the achieved overlap score significantly surpassed the results of state-of-the-art methods. 

\begin{table*}[th!]
	\centering
	\begin{tabular}{|c|c|c|c|c|}
		\hline
		& InvertedNet & \begin{tabular}[c]{@{}c@{}}All-\\ Dropout\end{tabular} & \begin{tabular}[c]{@{}c@{}}All-\\ Convolutional\end{tabular} & \begin{tabular}[c]{@{}c@{}}Original \\ U-Net\end{tabular} \\ \hline
		InvertedNet       & $\infty$      & 0.23 / \bm{$<0.01$} / 0.16                                                   & \bm{$<0.01$} / \bm{$<0.01$} / 0.56                                                    & \bm{$<0.01$} / \bm{$<0.01$} / 0.91                                                  \\ \hline
		All-Dropout       & 0.23 / \bm{$<0.01$} / 0.16        & $\infty$                                                 & \bm{$<0.01$} / \bm{$<0.01$} / 0.002                                                      & \bm{$<0.01$} / \bm{$<0.01$} /  \bm{$<0.01$}                                                  \\ \hline
		All-Convolutional & \bm{$<0.01$} / \bm{$<0.01$} / 0.56        & \bm{$<0.01$} / \bm{$<0.01$} / \bm{$<0.01$}                                                 & $\infty$                                                      & 0.03 / 0.35 / 0.37                                                     \\ \hline
		Original U-Net    & \bm{$<0.01$} / \bm{$<0.01$} / 0.91   & \bm{$<0.01$} / \bm{$<0.01$} / \bm{$<0.01$}                                              & 0.03 / 0.35 / 0.37                                                       & $\infty$                                                    \\ \hline
	\end{tabular}
	\caption{The significance difference analysis of segmentation results using Wilcoxon signed-rank test for Jaccard scores on the test set. The p-values are given for lungs, clavicles and heart (separated by "slash" sign).}
	\label{table:significance}
\end{table*}




\subsection{Timing Performance}

We performed all our experiments on a PC with Intel(R) Xeon(R) CPU E5-2650 v3 @ 2.30 GHz CPU and GeForce GTX TitanX GPU with 12 GB of memory. Training time for \originalunet, \, \udropall \, and \uallconv \, models was 12.4, 13.1 and 14.5 hours respectively. Despite a lower number of parameters training of the \invertednet \, took longer - 33 hours. This happened due to a larger number of high resolution features than in the other considered architectures networks. Table \ref{table:execution_time} shows an overview of the proposed architectures with execution times for both CPU and GPU.


\begin{table}[h!]
	\centering
	\renewcommand{\arraystretch}{1.2}
	\begin{tabular}{|c|c|c|c|c|c|}
				\hline
				Architecture & \# of params & CPU (s) & GPU (s) & Size\\ \hline
				\invertednet   &  3 140 771     & 7.1   & 0.06  & 12 MB\\ \hline
				\uallconv    & 34 512 388    &  4.2   & 0.03  &  131 MB\\ \hline
				\udropall   &  31 377 988   &  4.1   & 0.03  &  119 MB\\ \hline 
	\end{tabular}
	\caption{Overview of the proposed architectures for~256 x 256 imaging resolution}
	\label{table:execution_time}
\end{table}

To the best of our knowledge, our method is the fastest multi-class segmentation approach for CXR images to date. This will be particularly beneficial in large clinical environments where hundreds or sometimes thousands of people are screened every day.

\subsection{Limitations}

The original resolution of the JSRT dataset is $2048 \times 2048$. The manual annotations in the SCR database were created by van Ginneken et al.~\cite{Ginneken2006} on the downsampled images at $1024 \times 1024$. In this paper we concentrated on the $256 \times 256$ imaging resolutions to make the results comparable with the state-of-the-art approaches. Training with the imaging resolution of $1024 \times 1024$ effectively is possible however it would require using multiple GPUs or reducing number of feature maps in the \originalunet. In the latter case it would make the results incomparable with the results achieved on other imaging resolutions because the architecture would then become different and the former multiple GPUs scenario is a part of our ongoing work.

\begin{figure}[h!]
	\centering
	\includegraphics[width=0.49\textwidth]{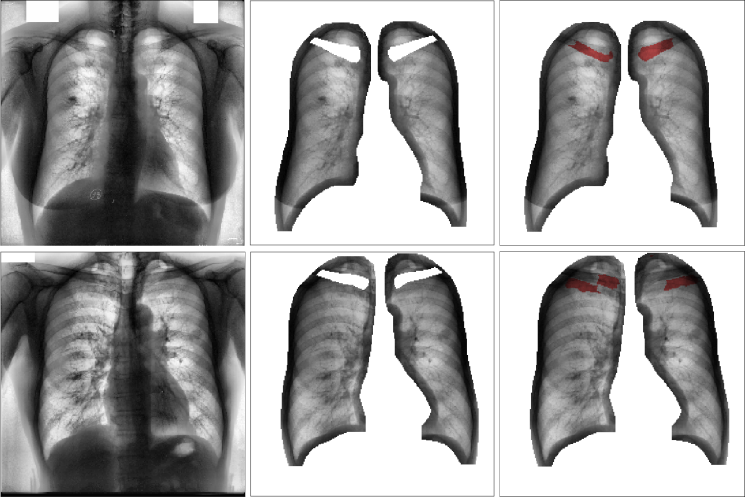}
	\caption{Two examples of clavicle segmentation task performed on the altered test images using the trained \emph{InvertedNet}: the original images (left) were clipped of everything but the lung fields and some padding around them (center). The network classifies pixels as clavicle based on the neighbourhood, indicating that local context is favoured over shape intrinsic information (right).}
	\label{fig:clavsLocal}
\end{figure}

Fully-convolutional architectures learn shape priors by a succession of learned feature detectors. The intrinsic set-up of such networks tries to find a trade-off between depth and computational feasibility. While increasing depth addresses larger semantic regions during processing, contraction elements such as max-pooling reduces the number of parameters but loses information about specific locations. As a result such network will favour local image context over shape, and by extension search and emphasize on boundaries over \emph{sound anatomy} in medical images. Fig.~\ref{fig:clavsLocal} aims to strengthen this intuition by visualizing the output of an image where everything but the lung fields has been clipped. The network, here \invertednet ,\, with all segmentation tasks included, still tries to classify the clavicles based on the local neighbourhood.

This indicates that the \emph{Dice-overlap} metric alone does not capture enough shape intrinsic information for plausible structures. Stricter regularization, e.g. introducing additional enforced adherence to shape priors as proposed by Oktay et al.~\cite{Oktay2017}, or favouring abstract loss objectives as described by Baumgartner et al.~\cite{Baumgartner2017}, may help in capturing semantically more meaningful segmentations. While potentially beneficial, especially in the case of InvertedNet where emphasis lies on the low level abstract features, this is beyond the scope of this paper, and subject to our future work.


\section{Conclusions}
\label{sec:conclusions}

This paper proposed an end-to-end approach for multi-class segmentation of anatomical organs in X-ray images. We introduced and evaluated three fully convolutional architectures which achieved high overlap test scores on the JSRT public dataset, matching or even outperforming the state-of-the-art methods on all considered organs. Our best architecture outperformed the human observer results for lungs and heart. To the best of our knowledge, our approach compares favourably to the state-of-the-art methods in the challenging clavicle segmentation task. Overall results show that adding more regularization and extracting larger numbers of high resolution low level abstract features can improve segmentation of smaller objects such as clavicles. Introducing weighting into the loss-function based either on cross entropy or Dice coefficients is important when dealing with the severe imbalanced data in CXR. Using exponential linear units can both speed up training and improve overall performance. We believe that with more training data and transfer learnt features from other CXR-related segmentation tasks, it is possible to further improve the scores. In future work we plan to extend the algorithm to other anatomical organs. 


More current research on the general domain aims at a deeper understanding of this kind of networks and thereby a deliberate enhancement in both architecture and training strategies. The core idea is to provide shorter connections between input and output layers, with the purpose of balancing different learning rates given the depth of a network, and thereby mitigating some of back-propagation problems. In case of our architectures, more research is needed to derive which branches can benefit of additional connections and which just add additional complexity. Ensemble networks, i.e. the simultaneous usage of multiple trained networks are gaining traction in current literature and competitions \cite{Ensemble2016, ImageNetWinners2017}. In the studied case of multi-class segmentation this potentially could find an intuitive application; nevertheless what explicit form this could have is a part of our future work. 

\ifCLASSOPTIONcaptionsoff
  \newpage
\fi

\bibliographystyle{IEEEtran}
\bibliography{references}


\end{document}